\documentclass{article} % For LaTeX2e
\usepackage{iclr2026_conference,times}
\iclrfinalcopy
\usepackage{amsmath,amsfonts,bm}

\def\eqref#1{equation~\ref{#1}}
\def\1{\bm{1}}

\DeclareMathAlphabet{\mathsfit}{\encodingdefault}{\sfdefault}{m}{sl}
\SetMathAlphabet{\mathsfit}{bold}{\encodingdefault}{\sfdefault}{bx}{n}

\usepackage{hyperref}
\usepackage{url}
\usepackage[utf8]{inputenc} % allow utf-8 input
\usepackage[T1]{fontenc}    % use 8-bit T1 fonts         % simple URL typesetting
\usepackage{booktabs}       % professional-quality tables
\usepackage{amsfonts}       % blackboard math symbols
\usepackage{nicefrac}       % compact symbols for 1/2, etc.
\usepackage{microtype}      % microtypography
\usepackage{xcolor}         % colors
\usepackage{algorithmic}
\usepackage[vlined, ruled]{algorithm2e}
\usepackage[utf8]{inputenc}
\usepackage{multirow}  % 用于多行合并单元格
\usepackage{booktabs}
\usepackage{graphicx}
\usepackage{microtype}
\usepackage{subcaption}
\usepackage{inconsolata}
\usepackage{times}
\usepackage{latexsym}
\usepackage[T1]{fontenc}
\usepackage[utf8]{inputenc}
\usepackage{microtype}
\usepackage{inconsolata}
\usepackage{graphicx}
\usepackage{colortbl}
\usepackage{amsmath}
\usepackage{booktabs}
\usepackage{amssymb}
\usepackage{adjustbox}
\usepackage{pifont}
\usepackage{bm}
\usepackage{longtable}
\usepackage{multirow}
\usepackage{multicol}
\usepackage[framemethod=TikZ]{mdframed}
\usepackage{xcolor}
\usepackage{url}
\usepackage{enumitem}
\usepackage{verbatim}
\usepackage{afterpage}
\usepackage{subcaption}
\usepackage{threeparttable}
\usepackage{array}
\usepackage{enumitem}
\usepackage{graphicx}
\usepackage{booktabs}  
\usepackage{hyperref}
\usepackage{mathtools}
\usepackage{amsthm}
\usepackage{siunitx}
\usepackage{amsfonts}
\usepackage{longtable}
\usepackage{xspace} % for smart spacing
\usepackage{listings}
\usepackage{wrapfig}
\usepackage{graphicx}

\title{MobileIPL: Enhancing Mobile Agents Thinking Process via Iterative Preference Learning}

\author{
    \textbf{Kun Huang$^{1*}$, Weikai Xu$^{21*\dag}$, Yuxuan Liu$^{31\dag}$, Quandong Wang$^{1}$, Pengzhi Gao$^{1}$, Wei Liu$^{1}$, } \\
    \textbf{Jian Luan$^{1}$, Bin Wang$^{1}$, Bo An$^{2\ddagger}$} \\
  $^1$\textmd{Xiaomi Inc.}
  $^2$\textmd{Nanyang Technological University}\\
  $^3$\textmd{Gaoling School of Artificial Intelligence, Renmin University of China} \\
  huangkun813@gmail.com, xuwk266@gmail.com \\
}

\begin{document}

\maketitle
\renewcommand{\thefootnote}{\fnsymbol{footnote}}
    \footnotetext[1]{Equal contribution.}
    \footnotetext[2]{Work done during the internship at XiaoMi. }
    \footnotetext[3]{Bo An is the corresponding author.}

\begin{abstract}
The Chain of Action-Planning Thoughts (CoaT) paradigm has been shown to improve the reasoning performance of VLM-based mobile agents in GUI tasks.
However, the scarcity of diverse CoaT trajectories limits the expressiveness and generalization ability of such agents. While self-training is commonly employed to address data scarcity, existing approaches either overlook the correctness of intermediate reasoning steps or depend on expensive process-level annotations to construct process reward models (PRM).
To address the above problems, we propose an \textbf{\underline{I}}terative \textbf{\underline{P}}reference \textbf{\underline{L}}earning (\textit{IPL}) that constructs a CoaT-tree through iterative sampling, scores leaf nodes using rule-based reward, and backpropagates feedback to derive Thinking-level Direct Preference Optimization (T-DPO) pairs. 
To prevent overfitting during warm-up supervised fine-tuning, we further introduce a three-stage instruction evolution, which leverages GPT-4o to generate diverse Q\&A pairs based on real mobile UI screenshots, enhancing both generality and layout understanding\footnote[4]{\url{https://huggingface.co/datasets/xwk123/MobileIPL-dataset}}.
Experiments on three standard Mobile GUI-agent benchmarks
demonstrate that our agent \textit{MobileIPL} outperforms strong baselines, including continual pretraining models such as OS-ATLAS and UI-TARS. It achieves state-of-the-art performance across three standard Mobile GUI-Agents benchmarks and shows strong generalization to out-of-domain scenarios.
\end{abstract}

\section{Introduction}
% v2.1

VLM-based mobile agents~\citep{wang2023enabling,ding2024mobileagent} have attracted considerable attention due to their ability to seamlessly interact with mobile graphical user interfaces (GUIs) and their potential to autonomously perform daily tasks.
Since actions are not directly specified in user instructions, mobile agents benefit from generating intermediate thoughts aligned with the current GUI context.
% Since actions cannot be directly inferred from user instructions, Mobile agents must reason about their decisions through intermediate thoughts grounded in the current GUI context. 
Recent work such as \textsc{AITZ}\citep{zhang2024android} has demonstrated that the Chain of Action-Planning Thoughts (CoaT) pattern---resembling the slow-thinking ``System 2'' process---is particularly effective in GUI domains.  

However, directly applying supervised fine-tuning (SFT) on CoaT trajectories may cause overfitting, leading the model to be trapped in fixed reasoning patterns. 
To address this limitation, recent studies in the general domain have explored self-training strategies. These approaches typically utilize the correctness of the final answer in output as a reward signal to train the model\citep{luong2024reft}. While effective in some contexts, relying solely on final answers overlooks the quality of intermediate reasoning steps, which can result in reward hacking and suboptimal reasoning processes.
% So recent studies in the general domain explore self-training strategies. For example, ReFT~\citep{luong2024reft} improves reasoning by evaluating the correctness of the final answer, but it overlooks the quality of the intermediate reasoning process, making it vulnerable to reward hacking. 
% ReST-MCTS~\citep{xie2024monte} addresses this by learning a process reward model (PRM) to score individual reasoning steps and perform tree-search policy to improve solution space quality, but this requires large-scale manual annotation of intermediate steps~\citep{guo2025deepseek-r1}. 
Some search-based approaches, such as ReST-MCTS~\citep{xie2024monte}, tackle this problem by learning a process reward model (PRM) to evaluate individual reasoning steps. 
% They further apply a tree-search policy to improve the quality of generated solutions. 
However, these approaches often require large-scale manual annotation of intermediate steps~\citep{guo2025deepseek-r1}. This challenge is especially severe in the Mobile GUI Agent domain. 
Unlike text-based tasks in coding or math, GUI environments rely on real devices or simulators, making step-level reward annotation significantly more costly and labor-intensive.
% Some search (eg ReST-MCTS\citep{xie2024monte}) approach addresse this problem by learning a process reward model (PRM) to evaluate individual reasoning steps. It further applies a tree-search policy to improve the quality of the generated solutions.However, this method requires large-scale manual annotation of intermediate steps~\citep{guo2025deepseek-r1}.This challenge is even more severe in the GUI-agent domain. Unlike text-based tasks in code or math, GUI environments depend on real devices or simulators, making the annotation of step-level rewards significantly more expensive and labor-intensive.

To address these limitations, we propose an iterative sampling framework that constructs a CoaT-tree based on Monte Carlo Tree Search (MCTS). 
% Rather than relying solely on final answers, our method assigns scores to individual reasoning steps, enabling the construction of thinking-level Direct Preference Optimization (T-DPO) pairs. 
Instead of relying on a PRM, we score each reasoning step and construct thinking-level DPO (T-DPO) pairs without manual step annotation.
Specifically, we perform multi-turn dialogue with a vision-language model (VLM) to incrementally build a CoaT-tree, where each node corresponds to a sampled response at a given reasoning step, conditioned on the dialogue history. This hierarchical structure captures diverse reasoning paths and facilitates fine-grained assessment of intermediate thoughts.
We first assign rewards to the leaf node, and then propagate these signals backward through the CoaT-tree to earlier reasoning steps. Based on the resulting values, we construct thinking-level DPO pairs to help agents optimize both final actions and the overall quality of their reasoning.
% Credit is assigned to intermediate nodes based on their contribution to high-reward trajectories, allowing us to optimize the model’s reasoning quality beyond the final action.

To mitigate the lack of diversity after warm-up SFT, we adopt an instruction evolution strategy. Specifically, we generate diverse Q\&A pairs grounded in real mobile UI screenshots from downstream training datasets. These Q\&A pairs serve two purposes: (1) prevent agents from overfitting to static downstream instructions by introducing varied reasoning contexts, and (2) improve agents' understanding of UI layouts through visually grounded question-answering.
We evaluate our approach on the CoaT dataset \textsc{AITZ} and long-horizon dataset AMEX, where it outperforms state-of-the-art GUI-agent continual pretraining agents such as OS-ATLAS~\citep{wu2024atlas} (+4.04\%) and UI-TARS~\citep{qin2025ui} (+3.54\%). Furthermore, experiments on the AndroidControl dataset demonstrate the strong generalization capability of our method to unseen apps and instructions (tasks). Under limited training resources, IPL consistently outperforms naive DPO using only half of the data for one iterative training round (+4.5\%), or one-fifth of the data for two iterative training rounds (+0.3\%). 
Analytical experiments show instruction evolution simultaneously improves both the diversity and quality of reasoning.

Overall, our main contributions are summarized as follows:

$\bullet$We propose an iterative framework to construct a CoaT-tree, and utilize rule-based rewards with backward credit assignment to form thinking-level DPO pairs for reasoning optimization.
% $\bullet$ We propose an interative framework to construct a CoaT-tree, which enables the generation of thinking-level Direct Preference Optimization (DPO) pairs. By performing backward credit assignment along the tree structure, we evaluate intermediate reasoning steps using a rule-based scoring function tailored to GUI tasks.

$\bullet$ We introduce an instruction evolution strategy to mitigate overfitting during warm-up SFT, enhancing the model's generalization and UI understanding.
% $\bullet$ We introduce an instruction evolution strategy to mitigate overfitting during warm-up supervised fine-tuning (SFT). 
% By leveraging GPT-4o to generate diverse QA pairs grounded in real mobile screenshots, our method enhances the model's ability to understand mobile UI layouts while maintaining generalization.

$\bullet$ We demonstrate the effectiveness of our method on three GUI-agent benchmarks: AITZ, AMEX, and AndroidControl. Furthermore, our approach even surpasses SoTA continual pretraining models.
\section{Related Work}
\subsection{Mobile GUI Agent}
LLMs~\citep{achiam2023gpt,xu2025llm,xu2025mobile,sun-etal-2024-determlr,liu2026come} are increasingly used as autonomous agents for mobile interaction~\citep{lieffects,wen2023empowering,liu2025mobilesteward,chen2025step}.
% LLMs \citep{achiam2023gpt} have emerged as autonomous agents \citep{lieffects,wen2023empowering} in the mobile domain and garnered considerable attention. % 这句不要
With the rapid development of vision-language models (VLMs), researchers build mobile GUI agents \citep{yang2023appagent,zheng2024gpt,qin2025ui,teamqwen2} and multi-agent frameworks \citep{ding2024mobileagent,li2024appagent,wang2024mobile,Luoling} based on closed-source VLMs. 
Meanwhile, some researchers focus on training agents with stronger element grounding \citep{cheng2024seeclick,wu2024atlas}, page navigation \citep{niu2024screenagent,lu2024gui,gou2024navigating,wang2025fedmobileagent}, GUI understanding \citep{you2024ferret,baechler2024screenai} and task planning capabilities \citep{zhang2024llamatouch,nong2024mobileflow,xu2024aguvis,qinghong2024showui,dorka2024training} based on open-source VLMs. 
Our method organizes trajectory data into multi-turns of dialogues based on the CoaT thinking pattern, preventing the agent becomes an action model with limited capabilities. 
\subsection{Reinforcement Learning}
The algorithms applied in natural language processing to align with human preferences include Direct Preference Optimization (DPO) \citep{rafailov2024direct}, Identity Preference Optimization (IPO) \citep{azar2024general}, Kahneman-Tversky Optimization (KTO) \citep{ethayarajh2023human}, and Proximal Policy Optimization (PPO) \citep{schulman2017proximal}. Specifically, ReFT \citep{luong2024reft} adopts reinforcement learning as a fine-tuning paradigm to improve performance on math problems.
% ReST-MCTS* \citep{zhang2024rest} focuses on a higher-quality process reward model for step quality. \citet{xie2024monte} labels the preference via MCTS based on feedback from self-evaluation.
% For GUI agents, \citet{bai2024digirl} and \citet{wang2024distrl} use joint online and offline reinforcement learning to improve the generalization of mobile agents and mitigate performance degradation when facing APP updates and unseen APPs.
% The RL algorithms used to align with human preferences include DPO 
% \citep{rafailov2024direct}, IPO \citep{azar2024general}, KTO \citep{ethayarajh2023human}, and PPO \citep{schulman2017proximal}. 
ReST-MCTS* \citep{zhang2024rest} focuses on the higher-quality step reward, where the process reward model is important. Xie, et al. \citep{xie2024monte} label the preference via MCTS based on feedback from self-evaluation.
For mobile GUI agents, Digirl \citep{bai2024digirl} and Distrl \citep{wang2024distrl} use online trajectory collection to improve the generalization of agents whose process is very slow. 
Reachagent \citep{wu2025reachagent} uses DPO training to compare the quality of multiple actions. 
% \bb{
TCPO \citep{jiao2025tcpo} also optimizes thoughts, but does not explicitly enforce thought–action consistency. TreePO \citep{li2025treepo}, TreeRL \citep{hou2025treerl}, and SPO \citep{guo2025segment} segment long sequences into many short segments, which leads to high computational cost and low data efficiency.
In contrast, our method models thoughts with a fixed CoaT-tree and uses T-DPO to optimize the thinking process, while step values are computed directly from rule-based rewards, without unstable PRMs. This design yields more efficient sampling and training, especially in GUI-agent settings.
% }
% Our method 将thought建模成了固定的CoAT-tree, uses T-DPO to optimize the thinking process, while the step value is calculated directly by rules without unstable PRMs.
\begin{figure*}[ht]
  \centering
  \includegraphics[width=0.85\textwidth]{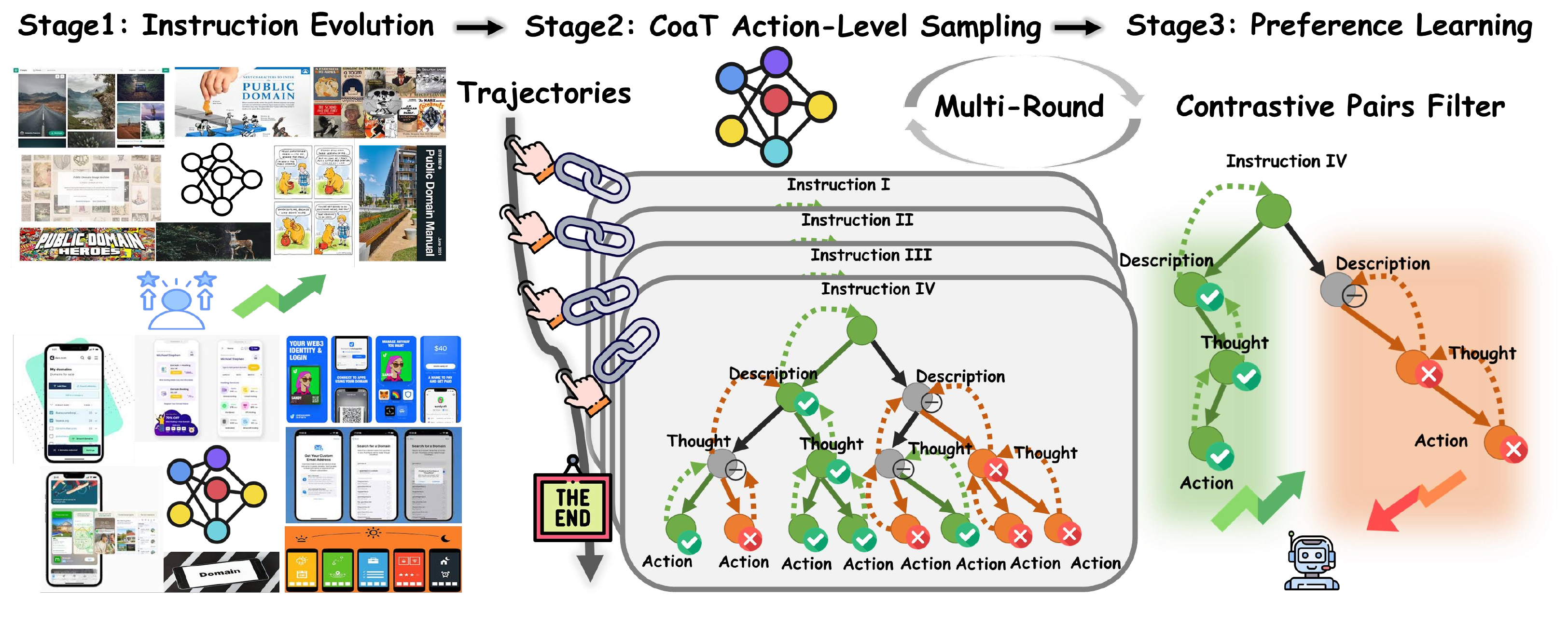}
  \caption{Overview of iterative preference learning framework. The left part presents the process of warm-up fine-tuning a general VLM to a mobile GUI domain agent with basic capabilities. The mid and right parts represent the iterative CoaT thinking-level sampling and T-DPO training process.}
  \label{fig:main}
\end{figure*}

\section{Methodology}
% As shown in Figure \ref{fig:main}, our method uses a three-stage instruction evolution to enhance the diversity of each round dialogue output, avoiding the agent focusing only on the thinking patterns contained in the CoaT data during warm-up SFT. 
% Initially, we supplement additional knowledge, diversify the rephrased instructions based on the existing training data through instruction evolution~(\S~\ref{3.1}), and use supervised learning to equip the model with basic capabilities~(\S~\ref{3.2}). Then, we perform sampling and rule selection at the CoaT level to construct contrastive data~(\S~\ref{3.3}). Finally, we optimize the model's thinking process through multiple rounds of Direct Preference Optimization~(\S~\ref{3.4}). 
% The detailed process of the entire method is presented in Algorithm \ref{alg: ipl_algorithm}.
In this section, we first introduce the multi-turn thinking process formulation ~(\S~\ref{3.1}) and explain our method.
% As shown in Figure \ref{fig:main}, our method starts with a three-stage instruction evolution strategy~(\S~\ref{3.2}). 
As shown in Figure \ref{fig:main}, our method starts with instruction evolution strategy~(\S~\ref{3.2}) to enhance output diversity in warm-up SFT stage. 
% This enhances the diversity of dialogue outputs in each round and prevents the agent from overfitting to the reasoning patterns in the CoaT data after warm-up SFT. Through this stage, the model acquires basic GUI-agent capabilities.
% ~(\S~\ref{3.2}).
Then, a CoaT-tree through iterative sampling~(\S~\ref{3.3}) is employed for each action. 
Every leaf node represents a complete action and is scored using a rule-based reward function. We then backpropagate the rewards along the tree to assign credit to intermediate reasoning steps. This process yields thinking-level contrastive pairs for DPO, which further improves the model’s reasoning ability. The detailed process is presented in Algorithm \ref{alg: ipl_algorithm}.

\subsection{Multi-turn Thinking Process Formulation} \label{3.1}
Each mobile GUI task contains a trajectory $\mathcal{T}$, several pages $u$, 
actions $\hat{a}$, and an instruction $I$, which can be represented as:
\begin{equation}
% \small
\mathcal{T} = \big\{I, u_0, \hat{a}_0, u_1, \hat{a}_1, \cdots, u_{n}, \hat{a}_{n}\}
\end{equation}
We formulate action $\hat{a}_{i}$ in the CoaT reasoning process as a multi-turn dialogue $\hat{a}_{i}=[s_1, s_2, s_3, s_4]$, where $s_i$ represents description, action-thought, action-decision, and grounding, respectively. This thinking paradigm based on the thinking–decision–grounding triplet, has been widely validated as effective in previous GUI works \citep{shen2024falcon,zhang2024android,qin2025ui,cheng2024seeclick}.
So the reasoning process can be formulated as:
\begin{equation}
% \small
s_1 = \mathrm{Description}  ( P_1, u_{i} )
\end{equation}
\begin{equation}
% \small
s_2 = \mathrm{Thought} ( P_2, u_{i}, I, \hat{a}_0,\cdots, \hat{a}_{i-1}, s_1 )
\end{equation}
$P$ represents each round of dialogue input prompt, $I$ is the task instruction, $u$ is the current GUI, and $\hat{a}_i$ is the step $i$ history action. 
Agents perform poorly when decoding the entire reasoning process in a single step, which is because image modal $u$ dominates the input tokens, surpassing textual instructions $I$ and action history $\hat{a}_i$, and diverting their attention away from the textual details. 
During autoregressive training, the agent is unaware that producing a final answer conforming to the required format is indispensable throughout the reasoning process.
Multi-turn thinking process effectively mitigates this problem, because additional dialogue steps guarantee a final answer is generated:
\begin{equation}
% \small
s_3 = \mathrm{Action} ( P_3, u_{i}, I, s_1, s_2 )
\end{equation}
\begin{equation}
% \small
s_4 = \mathrm{Grounding} ( P_4, u_{i}, I, s_1, s_2, s_3 )
\end{equation}
Previous work either performed RL in GUI-Agent directly on the trajectory without CoaT, missing the detailed thinking process of each action, or forced the model to bear the heavy burden of outputting the entire reasoning process at once.
In our method, when the reasoning process ends, the final $s_4$ is recorded as $\hat{a}_{n+1}$, step $i$ moves one step forward on the trajectory $\mathcal{T}$ and its thinking step reward is calculated recursively based on final step $s_4$. 
Dialogue-level textual input helps balance cross-modal token proportions and steers the agent’s attention toward the current reasoning step.
\subsection{Instruction Evolution}\label{3.2}
% Unlike mathematical problems, due to the lack of Mobile GUI-specific data during the pre-training stage, agents fail to generate diverse sampling thoughts like general domains. 
% % and leads to immediate overfitting on CoaT data after SFT with just a few epochs. 
% To address this limitation, we supplement the existing training screenshots in the trajectories $\mathcal{T}$ with annotated Q\&A through instruction evolution and diversify the rewritten instructions to generate the instruction evolution data $\mathcal{Q}$. 
% Specifically, as shown in Figure \ref{fig:ins}, the evolution process consists of three levels:
% \textbf{Level I: General GUI Q\&A tasks}. Grounding, Reference (Ref), and Page Descriptions are aimed at enhancing the agent's foundational capabilities. These tasks \citep{liu2024autoglm,yang2024aria} are proven to be the core capabilities of GUI agents during the pre-training.
\begin{wrapfigure}{r}{0.48\columnwidth}
  \setlength{\abovecaptionskip}{0.3cm}
  \setlength{\belowcaptionskip}{-0.5cm}
  \includegraphics[width=0.48\textwidth]{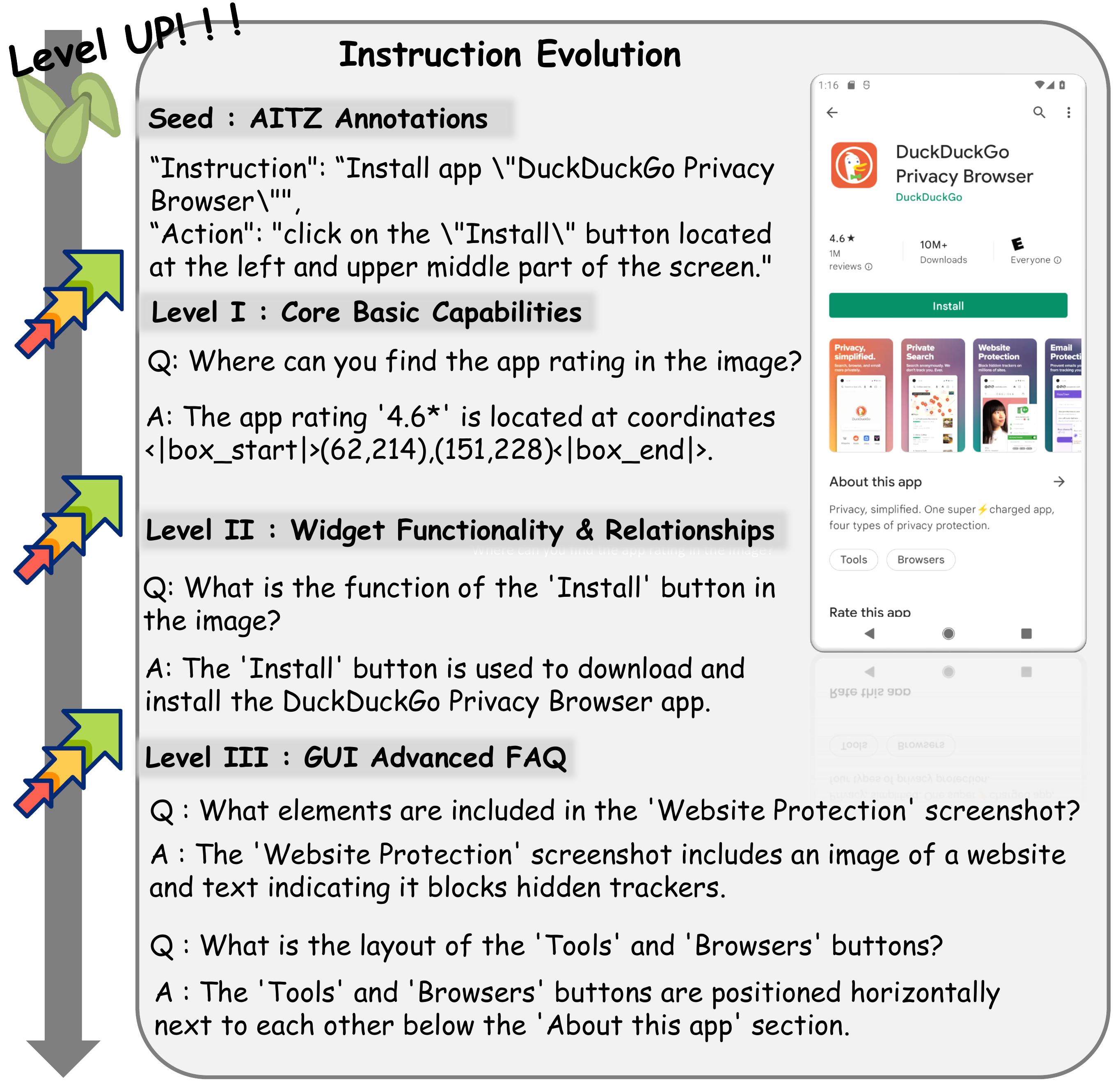}
  \caption{We process a three-stage instruction evolution and knowledge augmentation, enabling the agent to produce more diverse outputs for corresponding tasks while effectively mitigating overfitting.
  }
  \label{fig:ins}
\end{wrapfigure}
% As introduced in the previous section, the CoaT patterns in the mobile agent domain are often fixed. Agents tend to overfit on these fixed paradigms and struggle to generate diverse reasoning path after warm-up SFT training (in Sec \ref{sec:Output Space Sampling}). To address this challenge, we augment the original training trajectories $\mathcal{T}$ by attaching additional Q\&A annotations to UI screenshots through an instruction evolution process, resulting in a new dataset $\mathcal{Q}$ with more diverse instruction forms.
As discussed in the previous section, the CoaT patterns in the mobile agent domain are typically fixed. As a result, agents tend to overfit these static paradigms and struggle to generate diverse reasoning paths after the warm-up SFT training (as detailed in Sec \ref{sec:Output Space Sampling}). To address this issue, we enhance the original training trajectories, denoted as $\mathcal{T}$, by appending additional Q\&A annotations to UI screenshots through an instruction evolution process, thereby creating a new dataset $\mathcal{Q}$ with a broader range of instruction formats.
Specifically, as shown in Figure \ref{fig:ins}, the evolution process consists of three levels:

\noindent \textbf{Level I: General GUI Q\&A tasks}. Grounding, Reference (Ref), and Page Descriptions are aimed at enhancing the agent's foundational capabilities. These tasks \citep{liu2024autoglm,yang2024aria} are proven to be the core capabilities of GUI agents during the pre-training.

\noindent \textbf{Level II: Widget caption and relationships}. Descriptions of widget functions and the nested partition relationships between widgets. These tasks help agents understand the relationships between widgets, as previous work \citep{deng2024mobile} has found that agents tend to click on the textview, even in scenarios where the textview and the button are separate. 
% (\textit{e.g.,} ``In the Starbucks app, the beverage itself is separate from the \textit{`Add to Cart'} button.'').  
% Textview is used to describe the button, but is not interactive, while the button is the opposite.
% Understanding widget functions is crucial for the agent to complete page navigation tasks.

\noindent \textbf{Level III: GUI advanced FAQ}. Inspired by \cite{shen2024falcon}, we design an advanced FAQ that features more complex Q\&A, including descriptions of the page’s structural framework as well as expectations and predictions about navigation outcomes triggered by control interactions.
% and detailed information within the page. 
% (\textit{e.g.,} content about images within images).
% Rewriting is an essential method for increasing the diversity and robustness of agents, as CoaT action descriptions may refer to the same button using different expressions. (\textit{e.g., ``Click Install''} is equivalent to \textit{``Click the arrow block in the center of the screen''} or \textit{``Click the 'green' button.''})

\textbf{Warm-up Supervised Fine-tuning}:
To develop agents with standard thinking format and expand the reasoning space, we mix $\mathcal{T}$ and the instruction evolution data $\mathcal{Q}$, then perform warm-up SFT on  
$\mathcal D = \Big\{\mathcal{T},\mathcal{Q}\Big\}= \Big\{(u, e)^{(i)}\Big\}_{i=1}^{|\mathcal D|}$, 
where $u$ represents the prior knowledge (instructions, screenshot and action history) from $\mathcal{T}$ or the questions from $\mathcal{Q}$, and $e$ is the reasoning process from $\mathcal{T}$ or the answer from $\mathcal{Q}$ which is organized into multi-turn dialogues. 
To ensure output diversity, we select an earlier checkpoint with better potential correct space and diverse output to serve as the seed policy model. More details can be seen in Appendix \ref{appendix:seed_policy_model}. 
% \subsection{Warm-up Supervised Fine-tuning}\label{3.2}
% To develop an agent with basic task capabilities and expand the output sampling space, we mix $\mathcal{T}$ and the instruction evolution data $\mathcal{Q}$, then perform supervised fine-tuning (SFT) on  
% $\mathcal D = \Big\{\mathcal{T},\mathcal{Q}\Big\}= \Big\{(u, e)^{(i)}\Big\}_{i=1}^{|\mathcal D|}$, 
% where $u$ represents the prior knowledge (instructions, screenshot and action history) from $\mathcal{T}$ or the questions from $\mathcal{Q}$. $e$ is the reasoning process from $\mathcal{T}$ or the answer from $\mathcal{Q}$ which is organized into multi-round dialogues. The SFT loss can be computed as: 
% $$
% \mathcal L_{\mathrm SFT}(\theta) = -\mathbb E_{e \sim \mathcal D}[\log \pi_\theta(e|u)].
% $$
% To ensure output diversity, we select an earlier checkpoint with better potential correct space and diverse output to serve as the seed policy model. More details can be seen in Appendix 
% \ref{appendix:seed_policy_model}. 
% \subsection{CoaT Action Thinking Level Sampling}\label{3.3}
\subsection{Iterative Preference Learning}\label{3.3}
% After the warm-up SFT stage, agents build basic capabilities for GUIs and can sample outputs based on predefined inputs for each round of dialogue. 
% After warm-up SFT, the agent acquires basic GUI capabilities. We then perform iterative sampling to construct a CoaT-tree, where each path corresponds to a reasoning trajectory. To optimize the agent’s reasoning process, we form contrastive thinking-level DPO pairs using a rule-based reward.
% After warm-up SFT, the agent acquires basic GUI capabilities. 
% To further enrich the original thinking process, 
% We construct a CoaT-tree through iterative sampling for each reasoning step. Then assign the leaf node score based on rule-based reward function.Based on the score, construct thinking-level DPO pairs to optimize agents’ reasoning process.**
After the warm-up SFT, the agent acquires basic GUI capabilities. We construct a CoaT-tree by iteratively sampling each reasoning step and then assign a score to the leaf nodes based on a rule-based reward function. Using these scores, we generate thinking-level DPO pairs to optimize the agent's reasoning process.
% where each path represents a distinct reasoning trajectory. 
% Based on rule-based rewards, thinking-level DPO pairs are filtered to optimize agents’ reasoning process.

\noindent \textbf{Iterative Sampling \& Rule-based Reward.}
We iteratively sample each reasoning step along the CoaT paradigm \citep{zhang2024android}. 
% This allows us to build value functions for different tasks and reduces the pressure on the agent to output such long sentences at once. 
The $\mathcal{K}$ sampling results $(\hat{s}_{t}|\hat{s}_{1:t-1})^\mathcal{K}$ at step $t$ can be expressed as:
\begin{equation}
\hat{s}_t = \Big\{ (\hat{s}^{(k)}_{t} \mid \hat{s}_0, \cdots, \hat{s}_{t-1}) \Big\}^K_{k=1}
\end{equation}
Naturally, the final step in CoaT (the leaf node in the sampling tree) expresses a reward compared with the ground truth action $a^*$, which is then propagated back to other intra-nodes.
The formula for the rule-based reward of leaf nodes is as follows:
% \begin{equation}
% \small
% v(s_t) =
%     \begin{cases}
%     1,  & dis(s_t,a^*)_{[x,y]} < 0.05 \\
%     0.2, &  dis(s_t, a^*)_{[x, y]} \geq 0.5 \\
%     \frac{0.7 - dis(s_t, a^*)_{[x, y]}}{0.9}, & \\
%     &  0.05  < dis(s_t, a^*)_{[x, y]} < 0.5 \\
%     F_1, & F_1(s_t,a^*)_{text} > 0.2 \\
%     1, & dir(s_t \sim a^*)_{scroll} \\
%     0.1,  & type(s_t \sim a^*) \\
%     0,  &   else
%     \end{cases}
% \label{eq:leaf_value}
% \end{equation}
% \begin{equation}
% \small
% v(s_t) =
%     \begin{cases}
%     1,  & d(s_t,a^*)_{[x,y]} <= d_{min} \vee dir(s_t \sim a^*)_{scroll} \\
%      1 - (1-v_{format}) \cdot \frac{d(s_t, a^*)_{[x, y]}}{d_{max}-d_{min}}, & \\
%     &  d(s_t, a^*)_{[x, y]} >d_{min} \\
%     v_{type} + v_{format} + (1 - v_{type} - v_{format}) \cdot F_1, & F_1(s_t,a^*)_{text} \\
%     v_{type},  & type(s_t \sim a^*) \\
%     0,  &   else
%     \end{cases}
% \label{eq:leaf_value}
% \end{equation}
\begin{equation}
\small
v(s_t) =
    \begin{cases}
    1,  & s_t=a* \\
    v_{type} + score_{match},  & type(s_t \sim a^*) \\
    0,  &   others
    \end{cases}
\label{eq:leaf_value}
\end{equation}

\begin{equation}
\small
score_{match}=
\begin{cases}
     v_{format} + 1\cdot (1-d(x,y))-(v_{type}+v_{format})\cdot d(x,y), & type(a*)=CLICK\\
    v_{format} + (1 - v_{type} - v_{format}) \cdot F_1, & type(a*)=INPUT \\
    0, & others \\
\end{cases}
\end{equation}

The reward score $v(s_t)$ ranges from 0 to 1, with a fully correct prediction receiving a score of 1. 
% We use $v_{\text{type}}$, $v_{\text{format}}$ to indicate whether the predicted action type matches the ground truth and output format.
We use $v_{\text{type}}$ and $v_{\text{format}}$ to indicate whether the predicted action type and output format match the ground truth.
% For \texttt{click} and \texttt{Input} actions, we further evaluate their internal structure. 
% Specifically, we introduce $v_{\text{format}}$ to indicate whether essential arguments (e.g., coordinates or text) are correctly extracted. 
% The final reward is computed based on the similarity between the prediction and the ground truth:
% \begin{itemize}
% \item \textbf{Click:} A distance-based score between the predicted and ground-truth coordinates, scaled between $d_{\min}$ and $d_{\max}$.
% \item \textbf{Input:} The $F_1$ score between the predicted and ground-truth text.
% \end{itemize}
% The full reward function is defined in Equation~\ref{eq:leaf_value} and further discussed in Section~\ref{sec:Rule-based Reward Design}. 
% \bb{
For \texttt{click} and \texttt{input} actions, we further evaluate their internal structure using smooth rewards based on spatial distance $d(x,y)$ and text match $F_1$. The final reward is computed from the similarity between the prediction and the ground truth:
% }

\begin{itemize}
  \item \textbf{Click:} A distance-based score between the predicted and ground-truth coordinates, normalized to $[0,1]$; smaller distances yield higher scores.
  \item \textbf{Input:} The $F_1$ score between the predicted and ground-truth strings; greater textual overlap yields higher scores.
\end{itemize}

The full reward is defined in Equation~\ref{eq:leaf_value} and discussed further in Section~\ref{sec:Rule-based Reward Design}.

Based on the structure of the CoaT-tree, we recursively compute the value of each intermediate reasoning step. Specifically, the value of $s_{t-1}$ is computed as the average value of its $\mathcal{K}$ sampled continuations at $s_t$:
% As shown in Figure \ref{fig:sampling-case}, the value 1 is assigned for a strict action matched aligned with metrics (relative grounding box distance less than 0.05 ($d_{min}$), the $F_1$ score of input text beyond 0.5 or scroll direction is matched), 0 for unmatched, and other scores are scaled based on the deviation. 
% The recursive calculation formula for intra-nodes is as follows: 
\begin{equation}
v(s_{t-1}) = c\cdot \frac{1}{\mathcal{K}} \sum_{k=1}^\mathcal{K} v(s_{t}^{(k)})
\label{eq:intra_value}
\end{equation}
Here, $\mathcal{K}$ denotes the number of sampled continuations for each reasoning step, and $c$ is a discount factor. The parameter searching experiment for $\mathcal{K}$ is described in detail in Section \ref{para-searching}.

\noindent \textbf{Contrastive Data Filter.}
% \bb{
After obtaining the sampling tree and node values, we evaluate the quality of the trees and extract contrastive data.
We can divided the sampling trees into three categories $\mathcal{R} = \{\alpha, \beta, \gamma\}$ based on their output quality, and the classification standards of $\alpha, \beta, \gamma$ are as follows:
% }
\begin{equation}
\small{
\alpha = \frac{\Big| \big\{ \mathcal{S}^{(i)} \mid \forall v_k \in \mathcal{S}^{(i)}, v_k=1 \big\} \Big|}{\sum_{i=1}^{|\mathcal{T}|} |(u, e)^{(i)}|}
}
\end{equation}
\begin{equation}
\small{
\beta = \frac{\Big| \big\{ \mathcal{S}^{(i)} \mid \exists v_{k}, v_{k'} \in \mathcal{S}^{(i)}, v_{k}=1, v_{k'} \neq 1 \big\} \Big|}{\sum_{i=1}^{|\mathcal{T}|} |(u, e)^{(i)}|}
}
\end{equation}
\begin{equation}
\small{
\gamma = \frac{\Big| \big\{ \mathcal{S}^{(i)} \mid \forall v_k \in \mathcal{S}^{(i)}, v_k \neq 1 \big\} \Big|}{\sum_{i=1}^{|\mathcal{T}|} |(u, e)^{(i)}|}
}
\end{equation}
$\mathcal{S}^{(i)}$ and $v_k$ refer to the instruction $i$ sampling tree and the $k$-th leaf nodes value of $\mathcal{K}$ sampled output. 
$\alpha$ is considered a perfect sampling tree, which can stably output correct thoughts and actions with in-domain trajectories, $\beta$ represents potential correct trees that can be used to construct contrastive data, and $\gamma$ denotes sampling trees that require refinement.
${\beta + \gamma}$ is considered a valid sampling space.
In $\beta$, actions with a value of 1 and as many diverse action types as possible are extracted as positive samples. 
In $\gamma$, the final ground truth action $a^*$ is used as a positive sample, but the intermediate steps of CoaT are not provided, and the pairs can be represented as:
\begin{equation}
\begin{aligned}
\beta_{pairs} &= \langle\hat{s}^{(k)}_{t}\uparrow ,\hat{s}^{(k')}_{t}\downarrow \mid (\hat{s}_1, \dots, \hat{s}_{t-1}), \\
& ~~~~~~~~~~~~~~ v(\hat{s}^{(k)}_{t}) - v(\hat{s}^{(k')}_{t}) > 1 / \mathcal{K} \rangle
\end{aligned} \label{eq:dpo_pair1}
\end{equation}
\begin{align}
\gamma_{pairs} &= \langle a^*\uparrow,\hat{s}^{(k)}_{t}\downarrow \mid \hat{s}_1, \dots, \hat{s}_{t-1}\rangle
\label{eq:dpo_pair2}
\end{align}

% \bb{
\textbf{Thinking-level Direct Preference Optimization}.
% }
% \bb{
After CoaT thinking-level Iterative Sampling, several positive and negative example pairs are collected.
During this stage, the agent policy undergoes updates through the above data-pairs, SFT loss, and CoaT-DPO loss \citep{rafailov2024direct}.
Suppose the agent gets values to pair $\langle +, -\rangle$ at CoaT step $t$, which are named $s^+_{t}$ and $s^-_{t}$; we have the agent performing a comparison for these pairs based on the same thoughts  $s_{1:t-1}$, which can be calculated as:
% }
% \bb{
\begin{equation}
\begin{aligned}
\mathcal L_{\mathrm{T\mbox{-}DPO}}= - \mathbb E_{(s_{1:t-1}, s^-_{t},s^+_{t})\sim \mathcal T_s}\bigg[\log\sigma(\beta \log\frac{\pi_\theta(s^+_{t}|s_{1:t-1})}{\pi_{ref}(s^+_{t}|s_{1:t-1})} &\\
- \beta \log\frac{\pi_\theta(s^-_{t}|s_{1:t-1})}{\pi_{ref}(s^-_{t}|s_{1:t-1})} )\bigg]&,
\end{aligned}
\end{equation}
% }
% The final loss combines DPO and SFT losses:
% \begin{equation}
%     \mathcal L = \mathcal L_{\rm C\mbox{-}DPO} + c \cdot \mathcal L_{\rm SFT}
% \end{equation}
% \bb{
To further refine the agent’s performance post-optimization, we employ the updated agent as the new base agent to continue collecting contrastive CoaT-action level pairs for additional T-DPO training. 
This iterative process is maintained until the agent reaches the performance bottleneck.
% }

\begin{algorithm*}[ht!]
\setlength{\abovecaptionskip}{0.3cm}
\setlength{\belowcaptionskip}{-0.5cm}
\small
\caption{Iterative CoaT thinking-level sampling and DPO self-training.}
\renewcommand{\algorithmicrequire}{\textbf{Input:}}
\renewcommand{\algorithmicensure}{\textbf{Output:}}
\label{alg: ipl_algorithm}
\begin{algorithmic}[1]
% \DontPrintSemicolon
\REQUIRE {base VLM $\pi$, advanced annotated model $R_{SoTA}$, step-level trajectory data $\mathcal{T}$, instruction evolution Q\&A set $\mathcal{Q}$, number of sampling $\mathcal{K}$, golden action $a^*$, value function $v$, the sampled CoaT data $D$, number of iterations $\mathcal{N}$.}
\FOR {$i=1$ to $N_0$}
    \STATE $\mathcal{Q}^* \leftarrow$ instruction\_evolution($R_{\text{SoTA}}, \mathcal{T}$) \tcp{instruction evolution by GPT-4o}
    \STATE $\mathcal{Q} \leftarrow$ human\_evaluation($h, \mathcal{Q}^*$) \tcp{human filter}
\ENDFOR
\STATE $\pi_{S_0} \leftarrow $ Warm-up$\_$SFT($\pi, \mathcal{T}, \mathcal{Q}$) \tcp{fine-tune seed model}
\FOR {$n=1$ to $\mathcal{N}$}
    \FOR {$i=1$ to $|\mathcal{T}|$}
        \STATE $D_{i} \leftarrow $ generate$\_$sampling$\_$thought($\pi_{S_{n-1}}$, $\mathcal{T}_{i}$, $\mathcal{K}$) \tcp{CoaT Sampling}
        % \FOR {$q=1$ to $\mathcal{N}$}
        \STATE $V^{leaf}_{i} \leftarrow v(D_{i}, a^*_{i})$ \tcp{match and calculate leaf values using Eq(\ref{eq:leaf_value})}
        \STATE $V^{intra}_{i} \leftarrow $ recursive$\_$calculate($D_{i}, V^{leaf}_{i}$)
        \tcp{recursive intra node values using Eq(\ref{eq:intra_value})} 
        \STATE $ D^+_{i}, D^-_{i} \leftarrow$ contrastive\_data\_filter($D_{i}, V_{i}$) 
        \tcp{filter positive and negative data using Eq(\ref{eq:dpo_pair1}, \ref{eq:dpo_pair2})}
        % \STATE $ D^+_{pos_{i,j}} \leftarrow D_{G_{i}(A_q=a^*)|_{q=1}^\mathcal{N}}$
        % \tcp{filter positive data}
        % \STATE $ D^-_{neg_{i,j}} \leftarrow D_{G_{i}(A_q\neq a^*)|_{q=1}^\mathcal{N}}$
        % \tcp{filter negative data}
        % \ENDFOR
    \ENDFOR
\STATE $\pi_{S_n} \leftarrow $ DPO($\pi_{S_{k-1}}, D^+, D^-$) \tcp{DPO self-training reference model}
\ENDFOR
% \STATE \textbf{Return} $\pi_{S_T}, V_{T}$
\ENSURE $\pi_{S}, D_G, \mathcal{Q}$
\end{algorithmic}
% \vspace{-5pt}
\end{algorithm*}
\vspace{-9pt}
\section{Experiments}
\subsection{Experiments Setups}

\noindent \textbf{Dataset.}
\textbf{\textsc{AITZ}} \citep{zhang2024android} is a high-quality trajectory set filtered and re-annotated from AITW \citep{rawles2024androidinthewild}, containing four subsets
% : general, install, single, and Google apps
, which also includes five types of actions
% : click, input, scroll, press, and stop.
% It is the first CoaT dataset for the mobile GUI domain, where the CoaT's description, thought, action, grounding, and expectations align with the general agent's standard perception, thought, decision-making, tool call, and reflection. 
\textbf{AMEX} \citep{chai2024amex} uses the same apps and action space as \textsc{AITZ}, but its task instructions are more complex and detailed, with an average trajectory length of 15+.
% \textbf{AndroidControl} \citep{li2024effects} includes different task types compared to the previous two datasets. 
\textbf{AndroidControl} \citep{li2024effects} includes OOD datasets, such as app unseen and task unseen.
% In addition, it contains extra actions such as wait and completely unseen out-of-domain test splits. 

\noindent \textbf{Metrics.} For evaluation, we use \textbf{Step.Acc} as metrics, consistent with Auto-GUI\citep{zhang2023you}, measures the agent's performance and uses \textbf{Action Type} to assess the degree of action type matching. 
This metric effectively evaluates the model's planning ability. 
% We use \textbf{Episodic}, which indicates the relative position where the first error occurs in the trajectory to bridge the gap between offline and online evaluation.

\noindent \textbf{Baselines.} Following prior work\citep{wu2024atlas}\citep{qin2025ui}, we use Qwen2-VL-7B~\citep{wang2024qwen2} as the backbone of our model.
We select CogAgent \citep{hong2024cogagent}, AUTO-GUI, Shpagent, OS-Atlas, UGround, UI-Tars and FedMobileAgent as baseline agents. 
GUI continuous pre-training agents can be further divided into two categories: (1) training the model as a GUI grounding agent, such as OS-Atlas-7B. (2) training the model as a general GUI agent, such as UI-Tars.
More details are provided in Appendix \ref{appendix:setup}.
\subsection{Main result}
\begin{table*}[t]
\centering
\caption{\textbf{Main results of \textbf{\textsc{AitZ}} dataset.} ZS, FT, PF, and IPL are short for zero-shot, fine-tuning, specific domain pre-training, and iterative preference learning, respectively. 
% For CLICK and TYPE actions, which are more complicated than the other three, we additionally report the action type prediction accuracy, marked as `type' in this table. 
% The total action-matching score is also included. 
% `GP' is short for goal progress. 
`-' represents that the agent or evaluation prompt is not open-sourced. 
Seed means the seed model for sampling and T-DPO training. 
$R_i$ refers to the number of iterations during training.
% The best result of each metric is marked in \textbf{bold} while the second is marked with \underline{underline}.
}

\renewcommand{\arraystretch}{1.2}
\resizebox{0.9\textwidth}{!}{
\begin{tabular}{ccccccccccc}
\toprule
\toprule
 \multirow{3}{*}{\textbf{Model}} & \multirow{3}{*}{\textbf{Mode}} & \multicolumn{9}{c}{\textbf{Atomic}}  \\ \cmidrule{3-11}
    &  & \multirow{2}{*}{\textbf{SCROLL}} & \multicolumn{2}{c}{\textbf{CLICK}} & \multicolumn{2}{c}{\textbf{TYPE}} & \multirow{2}{*}{\textbf{PRESS}} & \multirow{2}{*}{\textbf{STOP}} & \multicolumn{2}{c}{\textbf{Total}} \\ \cmidrule{4-7} \cmidrule{10-11}
    &  &   & \textbf{type}   & \textbf{match}   & \textbf{type}   & \textbf{match}  &  & & \textbf{type}   & \textbf{match}  \\ \midrule
  % CogAgent & ZS  & 56.41 & 79.90    & 51.50 & 67.40    & 34.00    & 48.30    & 4.76    & 65.86    & 44.52  \\ 
      CogAgent (CoaT) & ZS   & \underline{70.22} & 88.23    & 66.15 & 45.80    & 21.80    & 45.95    & 24.60   & 72.59   & 53.28  \\   
      % AUTO-GUI & FT  & \textbf{74.88} & 44.37    & 12.72 & 73.00    & 67.80    & 49.09    & 60.12   & 73.79    & 34.46  \\
      AUTO-GUI (CoaT) & FT    & 61.40 & 74.56    & 32.20 & 87.20   & 81.40    & 57.70    & 74.40   & \underline{82.98}    & 47.69    \\ 
     AriaUI-MoE & FT & 53.73 & 85.51 & 60.20 & 84.20 & 80.80 & 63.70 & 76.38 & 78.53 & 63.56 \\
    % \multicolumn{1}{l|}{Qwen2-VL-72B} & FT  & - & -    & - & -    & -    & -    & -  & 89.60 & 72.10  \\ 

  Seeclick-7B & PF & 11.14 & 69.92 & 52.96 & 53.80 & 53.00 & 67.88 & 55.36 & 62.93 &  49.11 \\
  UGround-7B & PF & 58.22 & 80.94 & 58.48 & 82.56 & 73.85 & 58.22   & 68.78  & 74.54 & 60.19  \\ 
  OS-Atlas-7B & PF & 76.12 & 75.82 & 54.83 & 87.80 &  81.60 & 68.67 & \textbf{81.75} & 77.83 & 65.11 \\
  UI-Tars-7B & PF & 52.50 & 83.03 & 64.27 & \textbf{89.97}  &  82.76  & 61.87 & 74.35 & 77.59 & 65.61 \\
  Falcon-UI-7B & PF & - & -    & - & -    & -    & -    & -  & \textbf{84.70} & 69.10 \\
 \midrule  
  \multicolumn{1}{l}{Qwen2-VL-7B (CoaT)} & FT & 47.50 & 81.53  & 59.72 & 81.96    & 73.85    & \textbf{58.22}    & 67.39   & 74.26    & 60.36  \\ 
 \midrule  
  AITZ-Seed & FT  & 42.83 & 82.48   & 53.16 & 82.56   &  75.29   & \underline{56.65}    & 61.82   & 73.14    & 55.40    \\
    % &  +R1   & 45.83 & \textbf{92.52}   & 71.12 & 87.77   &  81.23   & 23.49    & 73.55   & 78.74    & 65.36     \\ 
    % &  +R3    & 49.83 & \underline{91.64}   & \underline{71.01} & \underline{87.97}   & \underline{83.16}    & 49.86   & \underline{77.93}  & 81.37 & 68.62    \\ 
      MobileIPL & IPL  & 51.08 & \textbf{91.73}   & \textbf{71.45} &  \underline{88.20}   & \textbf{83.40}    & 51.69 & \underline{78.17}  & 81.90    & \textbf{69.15}    \\ 
\bottomrule
\bottomrule
\end{tabular}
}
\label{tab:aitz}
\vspace{-12pt}
\end{table*}

\noindent \textbf{\textsc{AITZ}.} 
As shown in Table \ref{tab:aitz}, 
% except for SCROLL and PRESS, 
MobileIPL achieves SoTA performance on most metrics. 
The reason for the lower PRESS Acc. is discussed in Section \ref{IPL Scaling} and Appendix \ref{annotation preferences}.
Multiple rounds of T-DPO improve MobileIPL by more than 10\% (55.40\% -> 69.15\%) compared to the seed model MobileIPL and Qwen2-VL-7B (60.36\% -> 69.15\%). 
% Additionally, due to the inaccuracy of widget bounding box annotations, MobileIPL converges after 10 rounds of DPO. 
Compared to continuous pre-training agents such as Falcon-UI, which is pre-trained on three million GUIs, MobileIPL still surpasses a performance difference of 0.05\%. 
The amount of training data required by our method is substantially smaller than that used by these pre-training approaches.
% T-DPO Round R1 yields the most significant improvement, increasing performance by nearly 10 points (55.40\% -> 65.36\%). 
% This demonstrates the extremely low training cost and strong effectiveness of our method. 
% The GUI pre-training Grounding model UGround underperforms its base model Qwen2-VL on \textsc{AITZ} downstream tasks (60.19\% < 60.36\%), but UI-Tars performs better (65.61\% > 60.36\%) because of its more varied pre-training tasks.

\begin{table*}[t]
  \centering
  \caption{\textbf{Main results on AMEX}. Seed means the seed model for sampling.
  % For example: ``In Gmail, please open the first email.''
  }
  \renewcommand{\arraystretch}{1.2}
  \resizebox{0.95\textwidth}{!}{%
  \begin{tabular}{cccccccccccc}
    \toprule
    \toprule
    \textbf{Model} & \textbf{Training Data} & \textbf{Gmail} & \textbf{Booking} & \textbf{Music} & \textbf{SHEIN} & \textbf{News} & \textbf{CM}	& \textbf{ToDo}	&  \textbf{Signal}	&  \textbf{Yelp} & \textbf{Overall}\\
    \midrule 
    SeeClick-7B &  $\textsc{AitW}+$External  &  28.2	 & 29.4	 &  18.1	 &  20.0	 & 30.0	 & 53.1 & 	30.7	 & 37.1	 & 27.4 & 	30.44 \\
    SphAgent-7B & \textsc{AitW} & 32.1	 & 45.9 &	46.1  &	35.1 &	48.3  &	61.1  &	55.9 &	43.3 &	42.9 &	45.63 \\ 
    % Ours & AMEX \\
    SphAgent-7B & AMEX &  61.7 & 68.2 & 77.7 & \textbf{72.0} & 71.9 & 64.6 & \textbf{79.6} & 71.3 & 69.6 & 70.71 \\
    AriaUI-MoE & AMEX & 63.1 & 62.3 & 68.5 & 58.9 & 83.0 & 54.7 & 62.5 & 83.3 & 66.9 & 64.10 \\
    UGround-7B & AMEX & 70.9 & 68.8 & 72.7 & 63.7 & 77.7 & 67.7 & 63.7 & 80.1 & 67.6 & 69.12  \\
    SphAgent-7B & \textsc{AitW} + AMEX & 62.4 & 	68.1 & 	76.3	 & 71.9	 & 68.6	 &  67.3 & 	77.6 & 	66.0	 & 64.1	  & 69.14 \\ 
    OS-Atlas-7B & AMEX &  61.1 & \textbf{73.5} & 77.9 & 61.6 & 75.2 & 66.4 & 71.0 & 75.9 & 72.0 & 70.33 \\
    UI-Tars-7B & AMEX & 67.7 & 70.0 & 71.8 & 63.8 & 71.5 & 67.7 & 77.0 & \textbf{86.4} &  72.8 &  70.33 \\
    \midrule
    \multirow{2}{*}{Qwen2-VL-7B~~~~} 
    % & AMEX & 45.9 & 64.5 & 74.4 & 71.8 &70.3 & 67.4 & 79.3 & 64.9 & 66.3 & 67.20 \\
    % & AMEX & 58.8 & 69.1 & 79.1 & 64.3 & 77.1 & 65.9 & 69.7 & 79.6 & \textbf{75.7} & 68.88 \\
    & AMEX & 58.0 & 70.1 & 76.6 & 63.8 & 79.4 & 66.8 & 67.8 & 80.2 & \textbf{76.6} & 69.01 \\
    & + CoaT & 75.9 & 68.1 & 77.7 & 66.2 & 76.8 & 66.4 & 77.5 & 79.6 & 65.6 & 70.93 \\
    % & + CoaT, Evol & 57.0 & 60.2 & 68.8 & 63.1 & 75.0 & 50.2 & 65.6 & 77.7 & 62.6 & 62.19 \\   
    \midrule
    \multirow{2}{*}{MobileIPL-7B~~~~} & AMEX (Seed) & 57.0 & 60.2 & 68.8 & 63.1 & 75.0 & 50.2 & 65.6 & 77.7 & 62.6 & 62.19 \\
    % & + R1 & 70.4 & 72.5 & 75.8 & 68.2 & 83.0 & 68.1 & 67.8 & 80.2 & 70.6 & 72.02 \\
    % & + R2 & 76.9 & 72.0 & 79.4 & 68.9 & 84.2 & \textbf{73.5} & 73.2 & \textbf{86.4} & 71.8 & 74.03 \\
    % & + R3 & 77.0 & \textbf{72.2} & 78.6 & 68.0 & \textbf{85.8} & 70.4 & 72.6 & 82.1 & 74.3 & 74.14 \\
    & MobileIPL & \textbf{77.3} & \underline{71.8} & \textbf{80.0} & 68.4 & \textbf{85.3} & \textbf{71.3} & 73.5 & 82.1 & 71.8 & \textbf{74.29} \\
    \bottomrule
    \bottomrule
    
  \end{tabular}
  }
  \label{tab:amex}
\vspace{-12pt}
\end{table*}

% \begin{wraptable}{r}{0.5\columnwidth}
% \centering
% \scriptsize
% \begin{tabular}{lccc}
% \toprule
% \textbf{Mode} & \textbf{Model} & \textbf{Grounding} & \textbf{Step.Acc} \\
% \midrule
% \multirow{2}{*}{ZS} 
%  & Claude* & 0.0 & 12.5  \\
%  & GPT-4o & 0.0 & 20.8 \\
% \midrule
% \multirow{3}{*}{FT} 
% & Aria-UI-7B & 43.2 & 10.2 \\
% & InternVL-2-4B & 72.7 & 66.7 \\
% & Qwen2-VL-7B & 68.5 & 69.1   \\
% \midrule
% \multirow{3}{*}{PF} 
% & OS-Atlas-7B & 78.5 & 71.2 \\ 
% & Falcon-UI-7B & - & 72.7  \\
% & UI-Tars-7B & \textbf{80.5} & 72.5  \\
% \midrule
% \multirow{3}{*}{IPL} 
% & + R1 & 74.5 & 72.0 \\
% & + R2 & 75.2 & 72.7 \\
% & + R3 & 75.7 & \textbf{73.6} \\
% \bottomrule
% \end{tabular}
% \caption{High-level instruction experiment results on \textbf{AndroidControl IDD}.}
% \label{tab:ac1}
% \end{wraptable}
% \begin{wraptable}{r}{0.5\columnwidth}
% \centering
% \caption{High-level instruction experiment results on \textbf{AndroidControl}.}
% \scriptsize
% \begin{tabular}{cccc}
% \toprule
% \textbf{Mode} & \textbf{Model} & \textbf{Grounding} & \textbf{Step.Acc} \\
% \midrule
% \multirow{2}{*}{ZS} 
%  & Claude* & 0.0 & 12.5  \\
%  & GPT-4o & 0.0 & 20.8 \\
% \midrule
% \multirow{3}{*}{FT} 
% & Aria-UI-7B & 43.2 & 10.2 \\
% & InternVL-2-4B & 72.7 & 66.7 \\
% & Qwen2-VL-7B & 68.5 & 69.1   \\
% \midrule
% \multirow{3}{*}{PF} 
% & OS-Atlas-7B & 78.5 & 71.2 \\ 
% & Falcon-UI-7B & - & 72.7  \\
% & UI-Tars-7B & \textbf{80.5} & 72.5  \\
% \midrule
% \multirow{1}{*}{Ours} 
% &  MobileIPL & 77.0 & \textbf{72.7} \\
% \bottomrule
% \end{tabular}
% \label{tab:ac1}
% \vspace{-12pt}
% \end{wraptable}
\noindent \textbf{AMEX.}
% As shown in Table \ref{tab:amex}, MobileIPL surpass the previous SOTA model SphAgent-B 3.58\%. And Mobile supass the baseline model (Qwen2-vl+Coat) 3.36\%. And Moible IPL surpass the OS-Atlas (+3.69\%) and UI-Tars(+3.69\%), which also used qwen2-vl as backbone. And with coat, the baseline model qwen2-vl increse  1.92\%,which means the CoaT patterns is effectiveness. 
% As shown in Table \ref{tab:amex}, MobileIPL surpasses the previous SOTA model, SphAgent-B, by 3.58\%. It also outperforms the baseline model (Qwen2-vl+Coat) by 3.36\%. Additionally, MobileIPL surpasses OS-Atlas (+3.69\%) and UI-Tars (+3.69\%), both of which also use Qwen2-vl as the backbone. With the incorporation of CoaT, the baseline model Qwen2-vl shows an increase of 1.92\%, demonstrating the effectiveness of CoaT patterns.
As shown in Table \ref{tab:amex}, MobileIPL surpasses the previous SOTA model, SphAgent-7B, by 3.58\%. It also outperforms the baseline model (Qwen2-vl+CoaT) by 3.36\%. Additionally, MobileIPL surpasses OS-Atlas (+3.69\%) and UI-Tars (+3.69\%), both of which also use Qwen2-vl as the backbone. With the incorporation of CoaT, the baseline model Qwen2-vl shows an increase of 1.92\%, demonstrating the effectiveness of CoaT patterns. In summary, these results confirm that MobileIPL delivers significant improvements over existing models in long trajectory scenarios.
\begin{table}[htbp]
% \vspace{-20pt}
\centering
\begin{minipage}[t]{0.48\textwidth}
\centering
\caption{High-level instruction experiment results on \textbf{AndroidControl}.}
\scriptsize
\begin{tabular}{cccc}
\toprule
\textbf{Mode} & \textbf{Model} & \textbf{Grounding} & \textbf{Step.Acc} \\
% \midrule
% \multirow{1}{*}{ZS} 
%  % & Claude* & 0.0 & 12.5  \\
%  & GPT-4o & 0.0 & 20.8 \\
\midrule
\multirow{3}{*}{FT} 
& Aria-UI-7B & 43.2 & 10.2 \\
& InternVL-2-4B & 72.7 & 66.7 \\
& Qwen2-VL-7B (SFT) & 68.5 & 69.1   \\
\midrule
\multirow{3}{*}{PF} 
& OS-Atlas-7B & 78.5 & 71.2 \\ 
& Falcon-UI-7B & - & 72.7  \\
& UI-Tars-7B & \textbf{80.5} & 72.5  \\
\midrule
\multirow{1}{*}{RL} 
& Qwen2-VL-7B(GRPO) & 70.7 & 69.8\\ 
\midrule
\multirow{1}{*}{Ours} 
&  MobileIPL & 77.0 & \textbf{72.7} \\
\bottomrule
\end{tabular}
\label{tab:ac1}
\end{minipage}
\hfill
\begin{minipage}[t]{0.48\textwidth}
\centering
\caption{High-level instruction results on \textbf{AndroidControl in-domain} and \textbf{OOD} subsets.}
\scriptsize
\begin{tabular}{ccccc}
\toprule
\textbf{Mode} & \textbf{Model} & \textbf{IDD} & \textbf{app-UN} & \textbf{task-UN} \\
\midrule
\multirow{3}{*}{FT} 
& PaLM 2S(full) & 65.5 & 58.7 & 59.7 \\
& PaLM 2S(LoRA) & 70.8 & 58.5 & 59.6 \\
& Qwen2-VL-7B(SFT) & 69.1 & 61.4 & 64.1 \\
\midrule
\multirow{3}{*}{PF} 
& FedMobileAgent & 54.7 & 52.3 & 51.2 \\
& SphAgent-7B & 69.4 & 57.1 & 62.9 \\
& OS-Atlas-7B & 71.2 & 60.7 & 66.2 \\
\midrule
\multirow{1}{*}{RL} 
& Qwen2-VL-7B(GRPO) & 70.2 & 68.1 & 69.7 \\
\midrule
\multirow{1}{*}{IPL} 
& MobileIPL-7B & \textbf{73.6} & \textbf{70.0} & \textbf{72.2} \\
\bottomrule
\end{tabular}
\label{tab:ac2}
\end{minipage}
% \vspace{-12pt}
\end{table}

\noindent \textbf{AndroidControl.}
As shown in Table \ref{tab:ac1}, MobileIPL achieves SOTA performance in Step.Acc (72.7\%), reaching the SOTA model Falcon-UI with fewer data. MobileIPL also outperforms continual pre-training agents in the GUI domain, such as OS-Atlas (+1.5\%) and UI-Tars (+0.2\%). Compared to the baseline model Qwen2-VL(SFT), MobileIPL not only improves Mobile Agent performance but also enhances grounding by 8.5\%. As shown in Table \ref{tab:ac2}, MobileIPL continues to achieve SOTA performance in unseen OOD settings, demonstrating strong generalization. In contrast, compared to performance in the IDD domain, the pre-trained model OS-Atlas shows a significant drop. MobileIPL exhibits less performance degradation in out-of-domain settings. We also ran GRPO with Qwen2-VL under the same computational resources, and found OOD performance similar to MobileIPL, because both are self-training. However, MobileIPL still outperforms GRPO in all subsets.
\subsection{Ablation Study}
% The CoaT pattern relies on multi-round dialogue to be effective. 
% We also need to explore the potential impact of diverse positive examples in the tree sampling subset $\alpha$, as these examples are not utilized in our DPO training. 
To test the effectiveness of IPL and instruction evolution, we conducted ablation experiments. First, removing IPL and using only SFT caused performance to drop from 65.4\% to 60.4\%, compared to the first round of MobileIPL, highlighting the crucial role that IPL plays. Next, removing instruction evolution led to a 2.5\% drop in IPL performance in the first round. This occurs because, without evolution, the model generates fewer training samples (156{,}418 -> 113{,}239). And as shown in Figure \ref{fig:combined_fig} (a), without instruction evolution, the diversity of model outputs decreased, causing a drop in IPL performance. This further confirms that instruction evolution is crucial for improving IPL.
% Since our method incorporates Instruction Evolution (Evo), we conduct an ablation study to analyze the contribution of each component. As shown in Table \ref{tab:ablation1}, removing IPL leads to a significant drop in overall performance from 65.4\% to 60.4\%, demonstrating that IPL plays a crucial role. In contrast, removing Evo results in only a marginal performance decrease (-0.1\%), indicating that Evo does not directly enhance the model's performance.
% We also examine the effects of dialogue (Dia) and history (His) information. 
% Removing historical data leads to a notable decline in the Stop metric (-5.7\%), suggesting a strong link between stopping decisions and historical context. 
% Introducing image history under LORA fine-tuning improved performance by 1.9\%, as it helped the model make more informed decisions when multiple actions were possible. 
% \begin{wraptable}{r}{0.63\columnwidth}
\begin{table}[h]
% \vspace{-0.3cm}
\caption{Ablation study results on \textsc{AITZ}.}
\centering
% \scriptsize
\begin{tabular}{lccccc}
\toprule
\textbf{Model} & \textbf{Scroll} & \textbf{Click} & \textbf{Type} & \textbf{Press} & \textbf{Total} \\
% \midrule
% MobileIPL-R6 & 51.1 & 71.5 & 83.4 & 51.7 & 69.2 \\
% ~~~ - IPL & 46.9 & 59.4 & 78.6 & 55.4 & 60.4 (-8.8) \\
% ~~~ - IPL, Evo & 47.5 & 59.7 & 73.9 & 58.2 & 60.3 (-8.9) \\
% ~~~ - IPL, Evo, Dia. & 46.8 & 55.7 & 81.8 & 59.2 & 59.2 (-10)\\
% ~~~ - IPL, Evo, Dia., His. & 47.9 & 56.8 & 79.0 & 56.6 & 58.9 (-10.3)\\
\midrule
MobileIPL-R1 & 45.8 & 71.1 & 81.2 & 23.5 & 65.4 \\
~~~ - IPL & 46.9 & 59.4 & 78.6 & 55.4 & 60.4 (-5.0) \\
~~~ - Evo (R1) & 44.8 & 67.7 & 78.8 & 24.0 & 62.9 (-2.5) \\
~~~ - IPL Negative (R1) & 46.9 & 61.1 & 74.2 & 56.6 & 61.4 (-4.0) \\
~~~ - IPL + Naive DPO (R1) & 47.5 & 59.7 & 73.8 & 58.2 & 60.3 (-5.1) \\
~~~ - 1/2 training data (R1) & 42.9 & 68.3 & 79.0 & 43.8 & 64.8 (-0.6)\\
~~~ - 4/5 training data (R2) & 30.8 & 67.1 & 77.6 & 33.2 & 60.6 (-4.8) \\
\bottomrule
\end{tabular}
\label{tab:ablation1}
% \vspace{-0.3cm}
\end{table}

% \noindent \textbf{Standard Self Training.}

\noindent \textbf{Ablation Study.}
Additionally, we remove negative samples from IPL-R1, training the model using only fully correct samples. This results in a 4.0\% performance drop, suggesting that negative samples help the model learn how to reason rather than merely memorize (SFT). 
% Furthermore, compared to training on the entire trajectory with naive DPO, the model’s performance drops from 65.4\% to 60.3\%, which proves the effectiveness of the CoaT-tree sampling and thinking process optimization.
Furthermore, training on the entire trajectory with naive DPO reduces performance from 65.4\% to 60.3\%. Compared with SFT trained on CoaT tree positive data (–IPL Negative), naive DPO is still 1.1 \% lower, confirming the effectiveness of CoaT-tree sampling and the thinking-process optimization.

\noindent \textbf{Low Resource.}
We also perform low-resource on AITZ, sampling 1/2 and 1/5 of the training data. As shown in table \ref{tab:ablation1}, using only half of the data, the first round of IPL training already outperformed the best results achieved by the original CoAT-SFT (-IPL) and naive DPO training. Furthermore, when using only one-fifth of the data, the second round of IPL training surpassed the performance of CoAT-SFT (-IPL), demonstrating the effectiveness of our method even in low-resource scenarios.
\subsection{Discussion and Analysis} \label{sec:Output Space Sampling}

\noindent \textbf{Rollout Efficiency and Performance Trade-off.} We compare MobileIPL with GRPO and the MCTS-style baseline SPO-Chain\citep{guo2025segment}. Although MobileIPL requires more rollout sampling than GRPO, it achieves better accuracy. Compared with SPO-Chain, MobileIPL uses only about half as many rollouts per sentence ($\sim$27 vs. $\sim$54) while still obtaining a +1.12 accuracy gain. We also observe that SPO-Chain is sensitive to hyperparameters (e.g., temperature and cut-point probabilities). Its best performance is achieved with a higher temperature (1.4) to enlarge the exploration space, but this also produces longer sentences and slows down sampling. Overall, MobileIPL offers a better efficiency–performance trade-off and is more practical for GUI settings.

\begin{table}[htbp]
\centering
\caption{Comparison of RL methods in terms of accuracy and rollouts per sentence.}
\label{tab:rollouts_stepacc}
\begin{tabular}{lcc}
\toprule
\textbf{Model} & \textbf{Accuracy} & \textbf{Rollouts per sentence} \\
\midrule
MobileIPL (Ours)      & 69.15 & $\sim$27 \\
SPO-Chain & 68.03 & $\sim$54 \\
GRPO      & 66.29 & 8 \\
\bottomrule
\end{tabular}
\end{table}

\begin{figure*}[t]
\setlength{\abovecaptionskip}{0.3cm}
\setlength{\belowcaptionskip}{-0.3cm}
\centering
\includegraphics[width=0.95\textwidth]{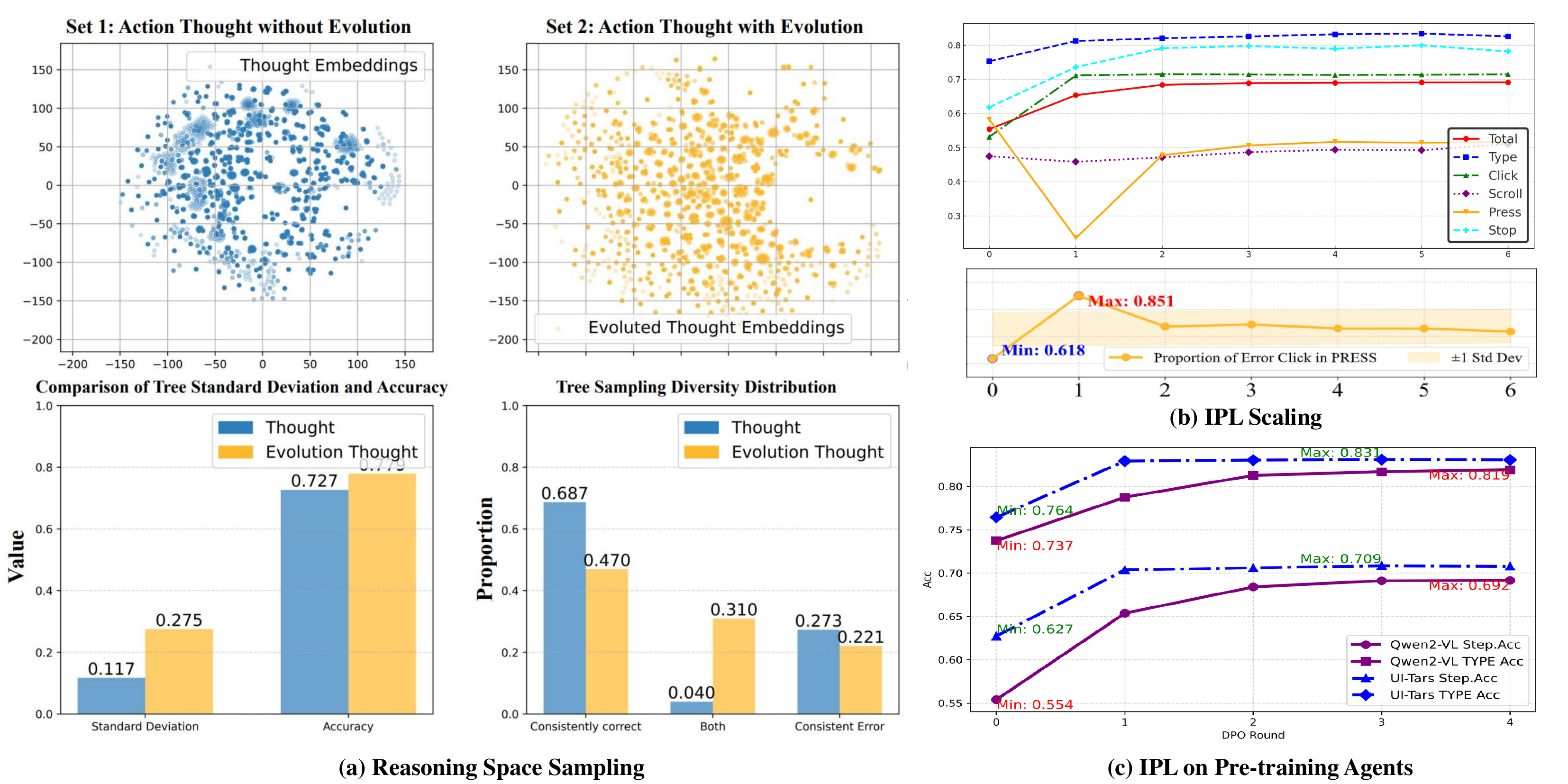}
\caption{
(a) Reasoning diversity before and after instruction evolution (left top) and the distribution of standard deviation and accuracy (left bottom).
(b) \textbf{Step.Acc} changes for action types in \textsc{AITZ} across IPL iterations (right top).
(c) The performance of UI-Tars-7B and Qwen2-VL-7B on \textsc{AITZ} as the seed model with 4-round IPL training (right bottom). }
\label{fig:combined_fig}
\end{figure*}

\noindent \textbf{Reasoning Space Sampling.} 
To evaluate the instruction evolution, we analyze the diversity of the sampling space for \textbf{Random 1000 steps}, the standard deviation of encoded embeddings, the dimensionality-reduced distribution, and the distribution of $S^{(i)}$ mentioned in Section \ref{3.3}. 
As shown in Figure \ref{fig:combined_fig} (a), the thoughts after instruction evolution exhibit a broader space than direct SFT.
Additionally, the embedding standard deviation within each tree increases significantly compared to the original data (+ 0.158). 
The diversified outputs do not negatively impact the agent's reasoning process, while the proportion of action sampling that includes the correct answer improves from 72.7\% to 77.9\%.
The bottom-right subplot reflects the distribution of output accuracy. 
\textbf{Consistently Correct} indicates that all samples for the current step match the golden answer, while \textbf{Consistently Error} is the opposite. 
\textbf{Both} represents cases where some samples are correct while others are incorrect, which serves as an ideal source for constructing T-DPO pairs. 
Compared to 47\% on the evolved data, the agent achieves 68.7\% convergence on the original data but exhibits a strong polarization(4\%). 
Three-stage instruction evolution significantly expands the sampling space (from 4\% to 31\%), proving that it simultaneously improves both the diversity and quality of reasoning. 
More details are in Appendix \ref{appendix:sampling}.

% \begin{wraptable}{r}{0.66\textwidth}
\begin{table}
% \vspace{-10pt}
\small
\centering
\caption{\textbf{Parameters Searching} on AITZ for the first round. $\mathcal{K}$ is the sampling number and R is the round of T-DPO learning.}
% \resizebox{0.65\textwidth}{!}
{
\begin{tabular}{cccccccc}
\toprule
\textbf{Parameter} & \textbf{TYPE} & \textbf{CLICK} & \textbf{SCROLL} & \textbf{PRESS} & \textbf{STOP} & \textbf{Total} \\
\midrule
$\mathcal{K}=2$ & 79.2 & 68.9 & 35.1 & 38.3 & 76.1 & 64.0 \\
$\mathcal{K}=3$ & 81.2 & 71.1 & 45.8 & 23.4 & 73.5 & 65.3 (\textbf{+1.3}) \\
$\mathcal{K}=4$ & 81.2 & 70.4 & 51.2 & 35.5 & 66.8 & 65.9 (\textbf{+0.6}) \\
\midrule
$R=0$ & 75.7 & 53.3 & 43.7 & 58.2 & 63.2 & 57.5 \\
$R=1$ & 77.5 & 71.1 & 43.3 & 23.5 & 67.0 & 61.2 (\textbf{+3.7}) \\
$R=2$ & 80.5 & 71.1 & 47.0 & 31.1 & 67.6 & 64.1 (\textbf{+2.9}) \\
$R=3$ & 82.0 & 71.5 & 47.2 & 47.8 & 79.1 & 68.4 (\textbf{+4.3}) \\
$R=4$ & 82.6 & 71.5 & 51.1 & 51.7 & 78.2 & 69.2 (\textbf{+0.8}) \\
\bottomrule
\end{tabular}
}
\label{tab:sample_number}
% \vspace{-5pt}
\end{table}

\noindent \textbf{Parameters Searching.} \label{para-searching}
We conducted an ablation study on the impact of the sampling number ($\mathcal{K}$) per stage and iterative round number ($R$).
% and based on the trade-off between performance and computational cost, we chose 3 as the final setting. 
As shown in the table \ref{tab:sample_number}, increasing the number of samples generally leads to better model performance. However, since our framework adopts a tree structure, increasing the sampling number from 3 to 4 causes the minimum number of tree nodes to grow significantly from $3^3 = 27$ to $4^3 = 64$. Despite this sharp increase, the performance improvement is limited (less than 1\%). Therefore, we adopt a sampling number of 3 for the final experiments. Regarding the number of rounds, we observe that both IPL performance and the size of the self-training dataset converge after several iterations. We therefore select the convergence round as our default setting. Additional details on computational cost are provided in Appendix~\ref{compute_use}.

% \begin{wraptable}{r}{0.65\columnwidth}
% \vspace{-0.3cm}
% \centering
% \scriptsize
% \begin{tabular}{lcccccc}
% \toprule
% \textbf{Method} & \textbf{SCROLL} & \textbf{CLICK} & \textbf{TYPE} & \textbf{PRESS} & \textbf{STOP} & \textbf{Total} \\
% \midrule
% IPL-R1 (full) & 45.83 & 71.12 & 81.23 & 23.49 & 73.55 & 65.36 \\
% IPL-R1 (1/2)  & 42.92 & 68.38 & 79.00 & 43.86 & 73.41 & 64.82 \\
% IPL-R2 (1/5)  & 30.78 & 67.07 & 77.60 & 33.16 & 65.28 & 60.62 \\
% Standard DPO (full) & 49.17 & 59.39 & 80.16 & 48.30 & 85.69 & 62.19 \\
% CoaT-SFT (full) & 47.50 & 59.72 & 73.85 & 58.22 & 67.39 & 60.36 \\
% \bottomrule
% \end{tabular}
% \caption{Performance comparison with varying data sizes and methods.}
% \label{tab:data_ablation2}
% \vspace{-0.3cm}
% \end{wraptable}

\noindent \textbf{IPL Scaling.}
\label{IPL Scaling}
% Although the total Step.Acc increases with IPL iterations, not all action types follow this upward trend. 
% As shown in Figure \ref{fig:combined_fig} (b), from the seed model to the first round of IPL, the accuracy of PRESS drops significantly (58.22\% -> 23.49\%), while CLICK increases (53.26\% -> 71.12\%). Inspired by action equivalence matching in AndroidControl, we attempt to analyze the reasons behind this phenomenon. 
% In fact, multiple actions on the same GUI can result in an equivalent outcome. 
% For example, after entering a search query, PRESS ENTER and directly click on a dropdown suggestion with the same keyword both navigate to the same page. 
Although overall Step.Acc increases across IPL iterations, not all action types follow this trend. As shown in Figure~\ref{fig:combined_fig}(b), from the seed model to the first IPL round, PRESS accuracy drops sharply (58.22\% → 23.49\%), whereas CLICK rises (53.26\% → 71.12\%). In the second round, however, PRESS accuracy rebounds. This stems from the severe underrepresentation of PRESS actions early on: the proportion of PRESS samples in the preference data grows from 1.6\% (1 round) to 10.9\% (2 round) as training progresses. With greater reasoning diversity and more PRESS-related examples, the model gradually learns PRESS behaviors and recovers accuracy in later rounds.
% Similarly, clicking the on-screen BACK button and pressing the hardware PRESS BACK key both navigate back from the current page. 
% And we analyze the proportion of steps where the ground truth action is PRESS, but the model outputs a different action. Inspired by action equivalence matching in AndroidControl, we attempt to analyze the reasons behind this phenomenon.
% We find that when PRESS accuracy drops to 23.49\%, CLICK accounts for 85.10\% of the errors in the lower part of the figure. 
% As more positive PRESS pairs (14.5\% -> 18.7\%) and less CLICK data (43.3\% -> 37.4\%) are sampled in subsequent IPL iterations, this error proportion gradually decreases to 73.7\%, and PRESS Acc. increases to 47.78\%. 
% For detailed statistical data, please refer to Appendix \ref{dpo scaling}.

\noindent \textbf{Iterative Preference Learning On GUI Continuous Pre-training Agent.} % hk自己写
As discussed in the previous experimental analysis, continuous pre-training in the GUI domain provides the agents with a stronger base model. 
However, we still need to explore the compatibility between post-training IPL, instruction evolution, and pre-training.
As shown in Figure \ref{fig:combined_fig} (c), UI-Tars outperforms Qwen2-VL-7B in all training stages, demonstrating better performance during the instruction evolution phase (62.7\% > 55.4\%). 
After four rounds of IPL, UI-Tars Step.Acc improves by 1.4\% compared to MobileIPL (69.2\% -> 70.6\%). 
More importantly, UI-Tars nearly converges after the first round of IPL, significantly reducing the number of sampling and preference learning iterations, thereby keeping the computational cost of post-training within an acceptable range. 
\section{Conclusion}
% In this paper, we propose Mobile Iterative Preference Learning (\textbf{MobileIPL}), a self-training GUI agent framework with instruction evolution, interative sampling in CoaT-tree, and rule-based reward.
% % In this paper, we propose Mobile Iterative Preference Learning (\textbf{MobileIPL}), a self-training GUI agent framework with instruction evolution, CoaT action sampling, and value calculation.
% % a GUI agent self-training framework that includes instruction evolution, CoaT action level output sampling, and value calculation method. 
% We also propose two new datasets: AMEX-CoaT and AndroidControl-CoaT. 
% We extensively validate MobileIPL on AITZ, AMEX, and AndroidControl, demonstrating its effectiveness. Furthermore, the Continuous Pre-training Experiments confirm its mutual reinforcement with pre-training, leading to enhanced performance.
In this paper, we propose Mobile Iterative Preference Learning (\textbf{MobileIPL}), a self-training GUI agent framework that incorporates instruction evolution, iterative sampling in the CoaT-tree, and a rule-based reward. 
% We also introduce two new datasets: AMEX-CoaT and AndroidControl-CoaT. 
We extensively evaluate MobileIPL on the AITZ, AMEX, and AndroidControl benchmarks, demonstrating its effectiveness. Furthermore, MobileIPL exhibits strong generalization capabilities on the OOD subsets of AndroidControl. 
% Experiments show that instruction evolution does not directly improve downstream task performance, but it effectively enhances the diversity of the model's outputs.
Experiments show that instruction evolution increases output diversity, generates more training data in IPL, and thereby improves IPL performance.
% While instruction evolution does not directly improve performance on downstream tasks, it fosters greater diversity in the model’s outputs. 
Finally, Continuous Pre-training experiments confirm the mutual reinforcement between MobileIPL and pre-training, leading to enhanced performance.

% \subsubsection*{Author Contributions}
% If you'd like to, you may include  a section for author contributions as is done
% in many journals. This is optional and at the discretion of the authors.

% \subsubsection*{Acknowledgments}
% Use unnumbered third level headings for the acknowledgments. All
% acknowledgments, including those to funding agencies, go at the end of the paper.
\section{Ethics Statement}
We have rigorously refined our dataset to remove any elements that could compromise personal privacy, thereby guaranteeing the highest level of protection for individual data. Instruction evolution was completed by AI SoTA close-sourced VLM, to whom we paid the necessary compensation to ensure that the training data was not leaked. The human evaluation of our work was carried out through a meticulously randomized selection of IT professionals. This process ensured a gender-balanced and educationally diverse panel, reflecting a wide spectrum of perspectives and expertise.

\section{Reproducibility statement}
All models and datasets used in this paper are open-source. The full experimental setup is detailed in Appendix \ref{appendix:setup}. Unless noted, all experiments use the same settings. We describe compute resources in Appendix \ref{compute_use}. Overall, these practices make our results reproducible.

\bibliography{iclr2026_conference}

@article{achiam2023gpt,
 author = {Achiam, Josh and Adler, Steven and Agarwal, Sandhini and Ahmad, Lama and Akkaya, Ilge and Aleman, Florencia Leoni and Almeida, Diogo and Altenschmidt, Janko and Altman, Sam and Anadkat, Shyamal and others},
 journal = {ArXiv preprint},
 title = {GPT-4 technical report},
 url = {https://arxiv.org/abs/2303.08774},
 volume = {abs/2303.08774},
 year = {2023}
}

@inproceedings{lieffects,
 author = {Wei Li and
William E. Bishop and
Alice Li and
Christopher Rawles and
Folawiyo Campbell{-}Ajala and
Divya Tyamagundlu and
Oriana Riva},
 bibsource = {dblp computer science bibliography, https://dblp.org},
 booktitle = {Advances in Neural Information Processing Systems 38: Annual Conference
on Neural Information Processing Systems 2024, NeurIPS 2024, Vancouver,
BC, Canada, December 10 - 15, 2024},
 editor = {Amir Globersons and
Lester Mackey and
Danielle Belgrave and
Angela Fan and
Ulrich Paquet and
Jakub M. Tomczak and
Cheng Zhang},
 title = {On the Effects of Data Scale on {UI} Control Agents},
 url = {http://papers.nips.cc/paper\_files/paper/2024/hash/a79f3ef3b445fd4659f44648f7ea8ffd-Abstract-Datasets\_and\_Benchmarks\_Track.html},
 year = {2024}
}

@article{wen2023empowering,
 author = {Wen, Hao and Li, Yuanchun and Liu, Guohong and Zhao, Shanhui and Yu, Tao and Li, Toby Jia-Jun and Jiang, Shiqi and Liu, Yunhao and Zhang, Yaqin and Liu, Yunxin},
 journal = {ArXiv preprint},
 title = {Empowering llm to use smartphone for intelligent task automation},
 url = {https://arxiv.org/abs/2308.15272},
 volume = {abs/2308.15272},
 year = {2023}
}

@article{yang2023appagent,
 author = {Yang, Zhao and Liu, Jiaxuan and Han, Yucheng and Chen, Xin and Huang, Zebiao and Fu, Bin and Yu, Gang},
 journal = {ArXiv preprint},
 title = {{AppAgent}: Multimodal agents as smartphone users},
 url = {https://arxiv.org/abs/2312.13771},
 volume = {abs/2312.13771},
 year = {2023}
}

@inproceedings{zheng2024gpt,
 author = {Boyuan Zheng and
Boyu Gou and
Jihyung Kil and
Huan Sun and
Yu Su},
 bibsource = {dblp computer science bibliography, https://dblp.org},
 booktitle = {Forty-first International Conference on Machine Learning, {ICML} 2024,
Vienna, Austria, July 21-27, 2024},
 publisher = {OpenReview.net},
 title = {GPT-4V(ision) is a Generalist Web Agent, if Grounded},
 url = {https://openreview.net/forum?id=piecKJ2DlB},
 year = {2024}
}

@article{ding2024mobileagent,
 author = {Ding, Tinghe},
 journal = {ArXiv preprint},
 title = {{MobileAgent}: enhancing mobile control via human-machine interaction and SOP integration},
 url = {https://arxiv.org/abs/2401.04124},
 volume = {abs/2401.04124},
 year = {2024}
}

@article{li2024appagent,
 author = {Li, Yanda and Zhang, Chi and Yang, Wanqi and Fu, Bin and Cheng, Pei and Chen, Xin and Chen, Ling and Wei, Yunchao},
 journal = {ArXiv preprint},
 title = {{AppAgent-V2}: Advanced Agent for Flexible Mobile Interactions},
 url = {https://arxiv.org/abs/2408.11824},
 volume = {abs/2408.11824},
 year = {2024}
}

@article{cheng2024seeclick,
 author = {Cheng, Kanzhi and Sun, Qiushi and Chu, Yougang and Xu, Fangzhi and Li, Yantao and Zhang, Jianbing and Wu, Zhiyong},
 journal = {ArXiv preprint},
 title = {{SeeClick}: Harnessing gui grounding for advanced visual gui agents},
 url = {https://arxiv.org/abs/2401.10935},
 volume = {abs/2401.10935},
 year = {2024}
}

@inproceedings{hong2024cogagent,
 author = {Wenyi Hong and
Weihan Wang and
Qingsong Lv and
Jiazheng Xu and
Wenmeng Yu and
Junhui Ji and
Yan Wang and
Zihan Wang and
Yuxiao Dong and
Ming Ding and
Jie Tang},
 bibsource = {dblp computer science bibliography, https://dblp.org},
 booktitle = {{IEEE/CVF} Conference on Computer Vision and Pattern Recognition,
{CVPR} 2024, Seattle, WA, USA, June 16-22, 2024},
 doi = {10.1109/CVPR52733.2024.01354},
 pages = {14281--14290},
 publisher = {{IEEE}},
 title = {CogAgent: {A} Visual Language Model for {GUI} Agents},
 url = {https://doi.org/10.1109/CVPR52733.2024.01354},
 year = {2024}
}

@article{wu2024atlas,
 author = {Wu, Zhiyong and Wu, Zhenyu and Xu, Fangzhi and Wang, Yian and Sun, Qiushi and Jia, Chengyou and Cheng, Kanzhi and Ding, Zichen and Chen, Liheng and Liang, Paul Pu and others},
 journal = {ArXiv preprint},
 title = {{Os-Atlas}: A foundation action model for generalist gui agents},
 url = {https://arxiv.org/abs/2410.23218},
 volume = {abs/2410.23218},
 year = {2024}
}

@inproceedings{niu2024screenagent,
 author = {Runliang Niu and
Jindong Li and
Shiqi Wang and
Yali Fu and
Xiyu Hu and
Xueyuan Leng and
He Kong and
Yi Chang and
Qi Wang},
 bibsource = {dblp computer science bibliography, https://dblp.org},
 booktitle = {Proceedings of the Thirty-Third International Joint Conference on
Artificial Intelligence, {IJCAI} 2024, Jeju, South Korea, August 3-9,
2024},
 pages = {6433--6441},
 publisher = {ijcai.org},
 title = {ScreenAgent: {A} Vision Language Model-driven Computer Control Agent},
 url = {https://www.ijcai.org/proceedings/2024/711},
 year = {2024}
}

@article{lu2024gui,
 author = {Lu, Quanfeng and Shao, Wenqi and Liu, Zitao and Meng, Fanqing and Li, Boxuan and Chen, Botong and Huang, Siyuan and Zhang, Kaipeng and Qiao, Yu and Luo, Ping},
 journal = {ArXiv preprint},
 title = {{GUI Odyssey}: A Comprehensive Dataset for Cross-App GUI Navigation on Mobile Devices},
 url = {https://arxiv.org/abs/2406.08451},
 volume = {abs/2406.08451},
 year = {2024}
}

@article{gou2024navigating,
 author = {Gou, Boyu and Wang, Ruohan and Zheng, Boyuan and Xie, Yanan and Chang, Cheng and Shu, Yiheng and Sun, Huan and Su, Yu},
 journal = {ArXiv preprint},
 title = {Navigating the digital world as humans do: Universal visual grounding for gui agents},
 url = {https://arxiv.org/abs/2410.05243},
 volume = {abs/2410.05243},
 year = {2024}
}

@article{chai2024amex,
 author = {Chai, Yuxiang and Huang, Siyuan and Niu, Yazhe and Xiao, Han and Liu, Liang and Zhang, Dingyu and Gao, Peng and Ren, Shuai and Li, Hongsheng},
 journal = {ArXiv preprint},
 title = {{AMEX}: Android Multi-annotation Expo Dataset for Mobile GUI Agents},
 url = {https://arxiv.org/abs/2407.17490},
 volume = {abs/2407.17490},
 year = {2024}
}

@article{you2024ferret,
 author = {You, Keen and Zhang, Haotian and Schoop, Eldon and Weers, Floris and Swearngin, Amanda and Nichols, Jeffrey and Yang, Yinfei and Gan, Zhe},
 journal = {ArXiv preprint},
 title = {{Ferret-UI}: Grounded Mobile UI Understanding with Multimodal LLMs},
 url = {https://arxiv.org/abs/2404.05719},
 volume = {abs/2404.05719},
 year = {2024}
}

@inproceedings{baechler2024screenai,
 author = {Gilles Baechler and
Srinivas Sunkara and
Maria Wang and
Fedir Zubach and
Hassan Mansoor and
Vincent Etter and
Victor Carbune and
Jason Lin and
Jindong Chen and
Abhanshu Sharma},
 bibsource = {dblp computer science bibliography, https://dblp.org},
 booktitle = {Proceedings of the Thirty-Third International Joint Conference on
Artificial Intelligence, {IJCAI} 2024, Jeju, South Korea, August 3-9,
2024},
 pages = {3058--3068},
 publisher = {ijcai.org},
 title = {ScreenAI: {A} Vision-Language Model for {UI} and Infographics Understanding},
 url = {https://www.ijcai.org/proceedings/2024/339},
 year = {2024}
}

@article{zhang2024llamatouch,
 author = {Zhang, Li and Wang, Shihe and Jia, Xianqing and Zheng, Zhihan and Yan, Yunhe and Gao, Longxi and Li, Yuanchun and Xu, Mengwei},
 journal = {ArXiv preprint},
 title = {{LlamaTouch}: A Faithful and Scalable Testbed for Mobile UI Task Automation},
 url = {https://arxiv.org/abs/2404.16054},
 volume = {abs/2404.16054},
 year = {2024}
}

@inproceedings{wang2024mobile,
 author = {Junyang Wang and
Haiyang Xu and
Haitao Jia and
Xi Zhang and
Ming Yan and
Weizhou Shen and
Ji Zhang and
Fei Huang and
Jitao Sang},
 bibsource = {dblp computer science bibliography, https://dblp.org},
 booktitle = {Advances in Neural Information Processing Systems 38: Annual Conference
on Neural Information Processing Systems 2024, NeurIPS 2024, Vancouver,
BC, Canada, December 10 - 15, 2024},
 editor = {Amir Globersons and
Lester Mackey and
Danielle Belgrave and
Angela Fan and
Ulrich Paquet and
Jakub M. Tomczak and
Cheng Zhang},
 title = {Mobile-Agent-v2: Mobile Device Operation Assistant with Effective
Navigation via Multi-Agent Collaboration},
 url = {http://papers.nips.cc/paper\_files/paper/2024/hash/0520537ba799d375b8ff5523295c337a-Abstract-Conference.html},
 year = {2024}
}

@article{nong2024mobileflow,
 author = {Nong, Songqin and Zhu, Jiali and Wu, Rui and Jin, Jiongchao and Shan, Shuo and Huang, Xiutian and Xu, Wenhao},
 journal = {ArXiv preprint},
 title = {{MobileFlow}: A Multimodal LLM For Mobile GUI Agent},
 url = {https://arxiv.org/abs/2407.04346},
 volume = {abs/2407.04346},
 year = {2024}
}

@article{xu2024aguvis,
 author = {Xu, Yiheng and Wang, Zekun and Wang, Junli and Lu, Dunjie and Xie, Tianbao and Saha, Amrita and Sahoo, Doyen and Yu, Tao and Xiong, Caiming},
 journal = {ArXiv preprint},
 title = {{Aguvis}: Unified Pure Vision Agents for Autonomous GUI Interaction},
 url = {https://arxiv.org/abs/2412.04454},
 volume = {abs/2412.04454},
 year = {2024}
}

@inproceedings{bai2024digirl,
 author = {Hao Bai and
Yifei Zhou and
Jiayi Pan and
Mert Cemri and
Alane Suhr and
Sergey Levine and
Aviral Kumar},
 bibsource = {dblp computer science bibliography, https://dblp.org},
 booktitle = {Advances in Neural Information Processing Systems 38: Annual Conference
on Neural Information Processing Systems 2024, NeurIPS 2024, Vancouver,
BC, Canada, December 10 - 15, 2024},
 editor = {Amir Globersons and
Lester Mackey and
Danielle Belgrave and
Angela Fan and
Ulrich Paquet and
Jakub M. Tomczak and
Cheng Zhang},
 title = {DigiRL: Training In-The-Wild Device-Control Agents with Autonomous
Reinforcement Learning},
 url = {http://papers.nips.cc/paper\_files/paper/2024/hash/1704ddd0bb89f159dfe609b32c889995-Abstract-Conference.html},
 year = {2024}
}

@article{wang2024distrl,
 author = {Wang, Taiyi and Wu, Zhihao and Liu, Jianheng and Hao, Jianye and Wang, Jun and Shao, Kun},
 journal = {ArXiv preprint},
 title = {{DistRL}: An asynchronous distributed reinforcement learning framework for on-device control agents},
 url = {https://arxiv.org/abs/2410.14803},
 volume = {abs/2410.14803},
 year = {2024}
}

@article{qinghong2024showui,
 author = {Qinghong Lin, Kevin and Li, Linjie and Gao, Difei and Yang, Zhengyuan and Wu, Shiwei and Bai, Zechen and Lei, Weixian and Wang, Lijuan and Shou, Mike Zheng},
 journal = {arXiv e-prints},
 pages = {arXiv--2411},
 title = {{ShowUI}: One Vision-Language-Action Model for GUI Visual Agent},
 year = {2024}
}

@article{dorka2024training,
 author = {Dorka, Nicolai and Marecki, Janusz and Anwar, Ammar},
 journal = {ArXiv preprint},
 title = {Training a Vision Language Model as Smartphone Assistant},
 url = {https://arxiv.org/abs/2404.08755},
 volume = {abs/2404.08755},
 year = {2024}
}

@article{schulman2017proximal,
 author = {Schulman, John and Wolski, Filip and Dhariwal, Prafulla and Radford, Alec and Klimov, Oleg},
 journal = {ArXiv preprint},
 title = {Proximal policy optimization algorithms},
 url = {https://arxiv.org/abs/1707.06347},
 volume = {abs/1707.06347},
 year = {2017}
}

@inproceedings{rafailov2024direct,
 author = {Rafael Rafailov and
Archit Sharma and
Eric Mitchell and
Christopher D. Manning and
Stefano Ermon and
Chelsea Finn},
 bibsource = {dblp computer science bibliography, https://dblp.org},
 booktitle = {Advances in Neural Information Processing Systems 36: Annual Conference
on Neural Information Processing Systems 2023, NeurIPS 2023, New Orleans,
LA, USA, December 10 - 16, 2023},
 editor = {Alice Oh and
Tristan Naumann and
Amir Globerson and
Kate Saenko and
Moritz Hardt and
Sergey Levine},
 title = {Direct Preference Optimization: Your Language Model is Secretly a
Reward Model},
 url = {http://papers.nips.cc/paper\_files/paper/2023/hash/a85b405ed65c6477a4fe8302b5e06ce7-Abstract-Conference.html},
 year = {2023}
}

@inproceedings{azar2024general,
 author = {Mohammad Gheshlaghi Azar and
Zhaohan Daniel Guo and
Bilal Piot and
R{\'{e}}mi Munos and
Mark Rowland and
Michal Valko and
Daniele Calandriello},
 bibsource = {dblp computer science bibliography, https://dblp.org},
 booktitle = {International Conference on Artificial Intelligence and Statistics,
2-4 May 2024, Palau de Congressos, Valencia, Spain},
 editor = {Sanjoy Dasgupta and
Stephan Mandt and
Yingzhen Li},
 pages = {4447--4455},
 publisher = {{PMLR}},
 series = {Proceedings of Machine Learning Research},
 title = {A General Theoretical Paradigm to Understand Learning from Human Preferences},
 url = {https://proceedings.mlr.press/v238/gheshlaghi-azar24a.html},
 volume = {238},
 year = {2024}
}

@techreport{ethayarajh2023human,
 author = {Ethayarajh, Kawin and Xu, Winnie and Jurafsky, Dan and Kiela, Douwe},
 institution = {Technical report, Contextual AI},
 title = {Human-centered loss functions (halos)},
 year = {2023}
}

@article{luong2024reft,
 author = {Luong, Trung Quoc and Zhang, Xinbo and Jie, Zhanming and Sun, Peng and Jin, Xiaoran and Li, Hang},
 journal = {ArXiv preprint},
 title = {{ReFT}: Reasoning with reinforced fine-tuning},
 url = {https://arxiv.org/abs/2401.08967},
 volume = {abs/2401.08967},
 year = {2024}
}

@inproceedings{zhang2024rest,
 author = {Dan Zhang and
Sining Zhoubian and
Ziniu Hu and
Yisong Yue and
Yuxiao Dong and
Jie Tang},
 bibsource = {dblp computer science bibliography, https://dblp.org},
 booktitle = {Advances in Neural Information Processing Systems 38: Annual Conference
on Neural Information Processing Systems 2024, NeurIPS 2024, Vancouver,
BC, Canada, December 10 - 15, 2024},
 editor = {Amir Globersons and
Lester Mackey and
Danielle Belgrave and
Angela Fan and
Ulrich Paquet and
Jakub M. Tomczak and
Cheng Zhang},
 title = {ReST-MCTS*: {LLM} Self-Training via Process Reward Guided Tree Search},
 url = {http://papers.nips.cc/paper\_files/paper/2024/hash/76ec4dc30e9faaf0e4b6093eaa377218-Abstract-Conference.html},
 year = {2024}
}

@article{xie2024monte,
 author = {Xie, Yuxi and Goyal, Anirudh and Zheng, Wenyue and Kan, Min-Yen and Lillicrap, Timothy P. and Kawaguchi, Kenji and Shieh, Michael},
 journal = {ArXiv preprint},
 title = {Monte Carlo Tree Search Boosts Reasoning via Iterative Preference Learning},
 url = {https://arxiv.org/abs/2405.00451},
 volume = {abs/2405.00451},
 year = {2024}
}

@article{shen2024falcon,
 author = {Shen, Huawen and Liu, Chang and Li, Gengluo and Wang, Xinlong and Zhou, Yu and Ma, Can and Ji, Xiangyang},
 journal = {ArXiv preprint},
 title = {{Falcon-UI}: Understanding GUI Before Following User Instructions},
 url = {https://arxiv.org/abs/2412.09362},
 volume = {abs/2412.09362},
 year = {2024}
}

@article{zhang2024android,
 author = {Zhang, Jiwen and Wu, Jihao and Teng, Yihua and Liao, Minghui and Xu, Nuo and Xiao, Xiao and Wei, Zhongyu and Tang, Duyu},
 journal = {ArXiv preprint},
 title = {{Android in the Zoo}: Chain-of-action-thought for gui agents},
 url = {https://arxiv.org/abs/2403.02713},
 volume = {abs/2403.02713},
 year = {2024}
}

@inproceedings{wang2023enabling,
 author = {Bryan Wang and
Gang Li and
Yang Li},
 bibsource = {dblp computer science bibliography, https://dblp.org},
 booktitle = {Proceedings of the 2023 {CHI} Conference on Human Factors in Computing
Systems, {CHI} 2023, Hamburg, Germany, April 23-28, 2023},
 doi = {10.1145/3544548.3580895},
 editor = {Albrecht Schmidt and
Kaisa V{\"{a}}{\"{a}}n{\"{a}}nen and
Tesh Goyal and
Per Ola Kristensson and
Anicia Peters and
Stefanie Mueller and
Julie R. Williamson and
Max L. Wilson},
 pages = {432:1--432:17},
 publisher = {{ACM}},
 title = {Enabling Conversational Interaction with Mobile {UI} using Large Language
Models},
 url = {https://doi.org/10.1145/3544548.3580895},
 year = {2023}
}

@inproceedings{rawles2024androidinthewild,
 author = {Christopher Rawles and
Alice Li and
Daniel Rodriguez and
Oriana Riva and
Timothy P. Lillicrap},
 bibsource = {dblp computer science bibliography, https://dblp.org},
 booktitle = {Advances in Neural Information Processing Systems 36: Annual Conference
on Neural Information Processing Systems 2023, NeurIPS 2023, New Orleans,
LA, USA, December 10 - 16, 2023},
 editor = {Alice Oh and
Tristan Naumann and
Amir Globerson and
Kate Saenko and
Moritz Hardt and
Sergey Levine},
 title = {AndroidInTheWild: {A} Large-Scale Dataset For Android Device Control},
 url = {http://papers.nips.cc/paper\_files/paper/2023/hash/bbbb6308b402fe909c39dd29950c32e0-Abstract-Datasets\_and\_Benchmarks.html},
 year = {2023}
}

@inproceedings{li2024effects,
 author = {Wei Li and
William E. Bishop and
Alice Li and
Christopher Rawles and
Folawiyo Campbell{-}Ajala and
Divya Tyamagundlu and
Oriana Riva},
 bibsource = {dblp computer science bibliography, https://dblp.org},
 booktitle = {Advances in Neural Information Processing Systems 38: Annual Conference
on Neural Information Processing Systems 2024, NeurIPS 2024, Vancouver,
BC, Canada, December 10 - 15, 2024},
 editor = {Amir Globersons and
Lester Mackey and
Danielle Belgrave and
Angela Fan and
Ulrich Paquet and
Jakub M. Tomczak and
Cheng Zhang},
 title = {On the Effects of Data Scale on {UI} Control Agents},
 url = {http://papers.nips.cc/paper\_files/paper/2024/hash/a79f3ef3b445fd4659f44648f7ea8ffd-Abstract-Datasets\_and\_Benchmarks\_Track.html},
 year = {2024}
}

@article{deng2024mobile,
 author = {Deng, Shihan and Xu, Weikai and Sun, Hongda and Liu, Wei and Tan, Tao and Liu, Jianfeng and Li, Ang and Luan, Jian and Wang, Bin and Yan, Rui and others},
 journal = {ArXiv preprint},
 title = {{Mobile-Bench}: An evaluation benchmark for llm-based mobile agents},
 url = {https://arxiv.org/abs/2407.00993},
 volume = {abs/2407.00993},
 year = {2024}
}

@article{yang2024aria,
 author = {Yang, Yuhao and Wang, Yue and Li, Dongxu and Luo, Ziyang and Chen, Bei and Huang, Chao and Li, Junnan},
 journal = {ArXiv preprint},
 title = {{Aria-UI}: Visual Grounding for GUI Instructions},
 url = {https://arxiv.org/abs/2412.16256},
 volume = {abs/2412.16256},
 year = {2024}
}

@article{liu2024autoglm,
 author = {Liu, Xiao and Qin, Bo and Liang, Dongzhu and Dong, Guang and Lai, Hanyu and Zhang, Hanchen and Zhao, Hanlin and Iong, Iat Long and Sun, Jiadai and Wang, Jiaqi and others},
 journal = {ArXiv preprint},
 title = {{AutoGLM}: Autonomous foundation agents for guis},
 url = {https://arxiv.org/abs/2411.00820},
 volume = {abs/2411.00820},
 year = {2024}
}

@article{zhang2023you,
 author = {Zhang, Zhuosheng and Zhang, Aston},
 journal = {ArXiv preprint},
 title = {You only look at screens: Multimodal chain-of-action agents},
 url = {https://arxiv.org/abs/2309.11436},
 volume = {abs/2309.11436},
 year = {2023}
}

@article{wang2024qwen2,
 author = {Wang, Peng and Bai, Shuai and Tan, Sinan and Wang, Shijie and Fan, Zhihao and Bai, Jinze and Chen, Keqin and Liu, Xuejing and Wang, Jialin and Ge, Wenbin and others},
 journal = {ArXiv preprint},
 title = {Qwen2-vl: Enhancing vision-language model's perception of the world at any resolution},
 url = {https://arxiv.org/abs/2409.12191},
 volume = {abs/2409.12191},
 year = {2024}
}

@article{qin2025ui,
 author = {Qin, Yujia and Ye, Yining and Fang, Junjie and Wang, Haoming and Liang, Shihao and Tian, Shizuo and Zhang, Junda and Li, Jiahao and Li, Yunxin and Huang, Shijue and others},
 journal = {ArXiv preprint},
 title = {{UI-TARS}: Pioneering Automated GUI Interaction with Native Agents},
 url = {https://arxiv.org/abs/2501.12326},
 volume = {abs/2501.12326},
 year = {2025}
}

@article{teamqwen2,
 author = {Team, Q},
 journal = {URL https://qwenlm. github. io/blog/qwen2},
 title = {Qwen2. 5-vl, January 2025}
}

@article{wang2025fedmobileagent,
 author = {Wang, Wenhao and Yu, Zijie and Liu, William and Ye, Rui and Jin, Tian and Chen, Siheng and Wang, Yanfeng},
 journal = {ArXiv preprint},
 title = {{FedMobileAgent}: Training Mobile Agents Using Decentralized Self-Sourced Data from Diverse Users},
 url = {https://arxiv.org/abs/2502.02982},
 volume = {abs/2502.02982},
 year = {2025}
}

@article{wu2025reachagent,
 author = {Wu, Qinzhuo and Liu, Wei and Luan, Jian and Wang, Bin},
 journal = {ArXiv preprint},
 title = {{ReachAgent}: Enhancing Mobile Agent via Page Reaching and Operation},
 url = {https://arxiv.org/abs/2502.02955},
 volume = {abs/2502.02955},
 year = {2025}
}

@article{guo2025deepseek-r1,
 author = {Guo, Daya and Yang, Dejian and Zhang, Haowei and Song, Junxiao and Zhang, Ruoyu and Xu, Runxin and Zhu, Qihao and Ma, Shirong and Wang, Peiyi and Bi, Xiao and others},
 journal = {ArXiv preprint},
 title = {Deepseek-r1: Incentivizing reasoning capability in llms via reinforcement learning},
 url = {https://arxiv.org/abs/2501.12948},
 volume = {abs/2501.12948},
 year = {2025}
}

@inproceedings{jiao2025tcpo,
  title={TCPO: Thought-Centric Preference Optimization for Effective Embodied Decision-making},
  author={Jiao, Kechen and Fang, Zhirui and Liu, Jiahao and Li, Bei and Wang, Qifan and Liu, Xinyu and Ruan, Junhao and Qiao, Zhongjian and Zhu, Yifan and Xu, Yaxin and others},
  booktitle={Proceedings of the 2025 Conference on Empirical Methods in Natural Language Processing},
  pages={9585--9599},
  year={2025}
}

@article{hou2025treerl,
  title={TreeRL: LLM Reinforcement Learning with On-Policy Tree Search},
  author={Hou, Zhenyu and Hu, Ziniu and Li, Yujiang and Lu, Rui and Tang, Jie and Dong, Yuxiao},
  journal={arXiv preprint arXiv:2506.11902},
  year={2025}
}

@article{li2025treepo,
  title={Treepo: Bridging the gap of policy optimization and efficacy and inference efficiency with heuristic tree-based modeling},
  author={Li, Yizhi and Gu, Qingshui and Wen, Zhoufutu and Li, Ziniu and Xing, Tianshun and Guo, Shuyue and Zheng, Tianyu and Zhou, Xin and Qu, Xingwei and Zhou, Wangchunshu and others},
  journal={arXiv preprint arXiv:2508.17445},
  year={2025}
}

@article{guo2025segment,
  title={Segment policy optimization: Effective segment-level credit assignment in rl for large language models},
  author={Guo, Yiran and Xu, Lijie and Liu, Jie and Ye, Dan and Qiu, Shuang},
  journal={arXiv preprint arXiv:2505.23564},
  year={2025}
}

@article{xu2025mobile,
  title={Mobile-Bench-v2: A More Realistic and Comprehensive Benchmark for VLM-based Mobile Agents},
  author={Xu, Weikai and Jiang, Zhizheng and Liu, Yuxuan and Gao, Pengzhi and Liu, Wei and Luan, Jian and Li, Yuanchun and Liu, Yunxin and Wang, Bin and An, Bo},
  journal={arXiv preprint arXiv:2505.11891},
  year={2025}
}

@article{xu2025llm,
  title={LLM-Based Agents for Tool Learning: A Survey: W. Xu et al.},
  author={Xu, Weikai and Huang, Chengrui and Gao, Shen and Shang, Shuo},
  journal={Data Science and Engineering},
  pages={1--31},
  year={2025},
  publisher={Springer}
}

@inproceedings{liu2025mobilesteward,
  title={Mobilesteward: Integrating multiple app-oriented agents with self-evolution to automate cross-app instructions},
  author={Liu, Yuxuan and Sun, Hongda and Liu, Wei and Luan, Jian and Du, Bo and Yan, Rui},
  booktitle={Proceedings of the 31st ACM SIGKDD Conference on Knowledge Discovery and Data Mining V. 1},
  pages={883--893},
  year={2025}
}

@article{liu2026come,
  title={CoME: Empowering Channel-of-Mobile-Experts with Informative Hybrid-Capabilities Reasoning},
  author={Liu, Yuxuan and Xu, Weikai and Huang, Kun and Chen, Changyu and Zhao, Jiankun and Gao, Pengzhi and Liu, Wei and Luan, Jian and Shang, Shuo and Du, Bo and others},
  journal={arXiv preprint arXiv:2602.24142},
  year={2026}
}

@article{chen2025step,
  title={STEP: Success-Rate-Aware Trajectory-Efficient Policy Optimization},
  author={Chen, Yuhan and Liu, Yuxuan and Zhang, Long and Gao, Pengzhi and Luan, Jian and Liu, Wei},
  journal={arXiv preprint arXiv:2511.13091},
  year={2025}
}

@inproceedings{sun-etal-2024-determlr,
    title = "{D}eterm{LR}: Augmenting {LLM}-based Logical Reasoning from Indeterminacy to Determinacy",
    author = "Sun, Hongda  and
      Xu, Weikai  and
      Liu, Wei  and
      Luan, Jian  and
      Wang, Bin  and
      Shang, Shuo  and
      Wen, Ji-Rong  and
      Yan, Rui",
    editor = "Ku, Lun-Wei  and
      Martins, Andre  and
      Srikumar, Vivek",
    booktitle = "Proceedings of the 62nd Annual Meeting of the Association for Computational Linguistics (Volume 1: Long Papers)",
    month = aug,
    year = "2024",
    address = "Bangkok, Thailand",
    publisher = "Association for Computational Linguistics",
    url = "https://aclanthology.org/2024.acl-long.531/",
    doi = "10.18653/v1/2024.acl-long.531",
    pages = "9828--9862"
}

@inproceedings{Luoling,
author = {Tan, Tao and Tu, Quan and Wu, Songhao and Sun, Hongda and Cheng, Chuanqi and Xu, Weikai and Yan, Rui},
title = {Luoling: An Immersive Cross-Modal Interactive Poem Creation Factory through Multi-Agent Collaboration},
year = {2025},
isbn = {9798400713316},
publisher = {Association for Computing Machinery},
address = {New York, NY, USA},
url = {https://doi.org/10.1145/3701716.3715184},
doi = {10.1145/3701716.3715184},
booktitle = {Companion Proceedings of the ACM on Web Conference 2025},
pages = {2911–2914},
numpages = {4},
location = {Sydney NSW, Australia},
series = {WWW '25}
}
\bibliographystyle{iclr2026_conference}

\appendix
\section{CoaT Thinking Process}
The table \ref{tab:method_comparison} summarizes the CoT paradigms (inputs and outputs) used in prior related works.
CoaT paradigm for fine-tuning agents: The effectiveness of the approach in AITZ stems from the inclusion of extra screen descriptions as part of the input, along with joint output of Screen Context, Action Think, Action Target, and Action Result. In contrast, our experiments show that a stage-wise multi-turn dialogue output leads to better performance. In this setup, the model focuses on a single sub-task at each stage, which not only improves clarity but also encourages a simplified and deliberate reasoning process. This insight aligns with UI-TARS, which only requires the model to generate a brief thought during inference.
Small-scale agent framing: Even models with relatively small parameter sizes can benefit from task-decomposed downstream training. For instance, OS-ATLAS and Falcon-UI adopt a similar architecture using GPT-4o for textual description and OS-ATLAS-base as the grounding model. They fine-tune models separately on different downstream tasks, resulting in a collection of OS-ATLAS-pro models, each specialized for a specific sub-task.
Large-scale prompting-based frameworks: Larger models typically adopt a multi-agent framework to support a CoT-style reasoning process. For example, AppAgent v2 and Mobile-Agent-v2 both utilize a plan–action–reflection architecture to complete tasks. In our work, we adopt a stage-wise CoaT multi-turn dialogue format, where the model focuses on one sub-task at a time. This design enables us to move away from the dependence on extra screen description inputs, as seen in AITZ, while leveraging the description + grounding structure proposed in OS-ATLAS to form the final structure of the MobileIPL CoaT paradigm. 
\begin{table}[h]
\centering
\scriptsize
\caption{Comparison of methods and their corresponding CoT paradigms.}
\begin{tabular}{l>{\raggedright\arraybackslash}p{2.5cm}>{\raggedright\arraybackslash}p{2.5cm}>{\raggedright\arraybackslash}p{5cm}}
\toprule
\textbf{Method} & \textbf{Backbone} & \textbf{Input} & \textbf{Output (CoT Paradigm)} \\
\midrule
Android in the Zoo & One fine-tuned agent & Instruction + image + screen description & Screen Context → Action Think → Action Target → Action Result \\
UI-TARS & One fine-tuned agent & Instruction + image & Thought → Action \\
OS-ATLAS & One fine-tuned agent + one large agent & Instruction + image & Action Description + Action Model \\
Mobile-Agent-v2 & Multi-agent + prompt engineering & Instruction + image & Stage 1: Plan → Stage 2: Action → Stage 3: Reflection \\
MobileIPL & One fine-tuned agent & Instruction + image & Stage 1: Description → Stage 2: Plan → Stage 3: Action → Stage 4: Grounding \\
\bottomrule
\end{tabular}
\label{tab:method_comparison}
\end{table}
\section{Selection of Seed Policy Model}\label{appendix:seed_policy_model}

In our preliminary experimental exploration, we discovered that for the seed policy model, better performance in the SFT phase does not necessarily translate to a higher upper bound in the subsequent IPL phase. This is because as training progresses, the model’s output space becomes increasingly aligned with the training data, reducing its diversity in sampling. Consequently, for incorrect instances, the model tends to generate erroneous outputs regardless of the sampling attempts.
To address this, we propose a sampling-oriented selection method for the seed policy model, incorporating the following two evaluation metrics:

\noindent \textbf{Sampling Accuracy($Acc_S$)}, which requires the model to hit more correct actions $a$ in the sampled output space $\mathcal{S}$.
\begin{equation}
\small
    Acc_S= \frac{\sum_{i=1}^{|\mathcal{T}|} \Big| \big\{  e_j^{(i)} \mid a^{(i)} \sim e_j^{(i)},  e_j^{(i)}\in\mathcal{S}^{(i)}\big\} \Big|}{\sum_{i=1}^{|\mathcal{T}|} |\mathcal{S}^{(i)}|}
\end{equation}

\noindent \textbf{Sampling Diversity($Div_R$)}, which requires the model to have a more diverse and extensive sampling space.
Standard deviation calculation of a single sampled tree \(\mathcal Dev_{{S}^{(i)}}\):
\begin{equation}
\small
Dev_{{S}^{(i)}} = \frac{1}{T} \sum_{t=1}^{T} \mathrm{StdDev} \left( \mathbf{E}(\hat{s}^{(k)}_t) \mid k = 1, \dots, \mathcal{K} \right)
\end{equation}
Among them, \(\mathbf{E}(\hat{s}^{(k)}_t)\) represents the representation of the \(k\)th sample output of the \(t\)th step after the encoder. 
Calculation of the standard deviation of the set $Div_S$:
\begin{equation}
Div_R = \frac{1}{N} \sum_{i=1}^{N} Dev_{{S}^{(i)}}
\end{equation}
where \( N \) is the number of sampled trees in the set $ \mathcal{R} $.

\section{Rule-based Reward Design}
\label{sec:Rule-based Reward Design}
\noindent \textbf{Derivation Of The Value Function.}
Our value function incorporates hyperparameters inspired by ReFT and is also influenced by the sampling number used during IPL. We explain the rationale behind key parameter choices in our method, especially those in Eq.(\ref{eq:leaf_value}), Eq.(\ref{eq:dpo_pair1}), and Eq.(\ref{eq:dpo_pair2}):

\noindent \textbf{Strong Reward}: We follow the ReFT \citep{luong2024reft} score to define strong reward signals, assigning values of 1 and 0, corresponding to fully correct and completely incorrect reasoning paths, respectively. 
In ReFT, a supervision signal of 0.1 encourages the model to produce a final answer following the predefined format. In our approach, this signal is repurposed to reward action type matching. 
Meanwhile, an additional $v_{format}$ reward is introduced to encourage proper formatting of actions.

\noindent \textbf{Weak Reward}:
For input action, the value linearly increases from $v_{format} + v_{type}$ up to the strong reward level, with $v_{format}$ acting as a threshold to distinguish weak from type-correct reward.
For grounding actions, values range between $v_{format} + v_{type}$ and 1, too. A value of $v_{format}$ indicates minimal correctness (e.g., extractable coordinates), while 1 indicates a closest match with the golden action, suitable for DPO pairing. 
Except $v_{format}$ and $v_{type}$ serving as discrete supervision signals, all other value signals are maintained as continuous.
1 / $\mathcal{K}$ in Eq. (\ref{eq:dpo_pair1}) arises naturally from our hierarchical training structure. For example, if one child is incorrect (e.g., value drops from 1 to 0), the average value for the parent node decreases by ~0.33 when the sampling number is 3. Thus, 1 / $\mathcal{K}$ serves as a meaningful threshold to distinguish positive vs. negative examples in the CoAT tree.

\textbf{Text F1 vs.\ Text-Embedding Similarity: }
We replaced the F1\mbox{-}based reward with a BERT\mbox{-}based semantic reward and evaluated both variants.
As shown in Table~\ref{tab:reward-ablation}, the F1 reward outperforms the BERT embedding reward across all metrics, with the largest gain on \textit{TYPE  ACTION (acc)} (+4.53\%).
This aligns with the importance of exact keyword matching in GUI input, indicating that F1 is better suited than semantic similarity for reward design in mobile UI input scenarios.

\begin{table}[t]
\centering
\caption{Text F1 vs.\ text-embedding similarity: ablation on reward design. Best results are in \textbf{bold}.}
\label{tab:reward-ablation}
\begin{tabular}{ccccc}
\toprule
\textbf{Reward Type} & \textbf{TYPE (type)} & \textbf{TYPE (acc)} & \textbf{TOTAL (type)} & \textbf{TOTAL (acc)} \\
\midrule
BERT & 84.17 & 76.70 & 77.81 & 62.49 \\
F1   & \textbf{87.78} & \textbf{81.23} & \textbf{78.74} & \textbf{65.37} \\
\bottomrule
\end{tabular}
\end{table}

% \begin{figure*}[ht]
%   \centering
%   \includegraphics[width=0.98\textwidth]{img/porpotion_click.pdf}
%   \caption{The proportion of CLICK and PRESS actions selected as positive data during the first four rounds of IPL.}
%   \label{fig:porpotion_click}
% \end{figure*}
\begin{figure}[htbp]
    \centering
    % 第一张子图
    \begin{subfigure}{0.47\textwidth}
        \centering
        \includegraphics[width=\linewidth]{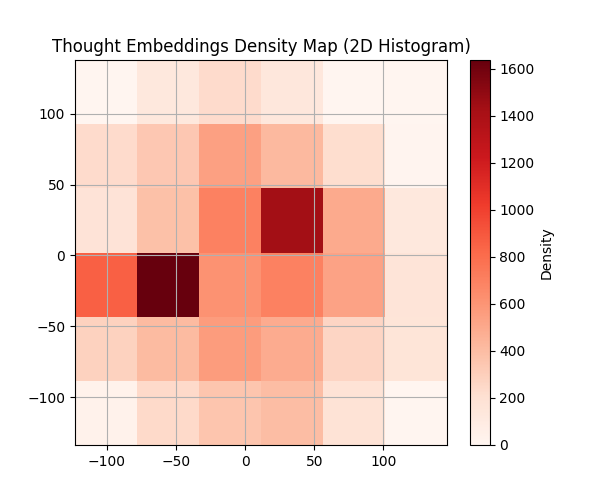}
        \label{fig:subheatmap1}
    \end{subfigure}
    \hfill
    % 第二张子图
    \begin{subfigure}{0.47\textwidth}
        \centering
        \includegraphics[width=\linewidth]{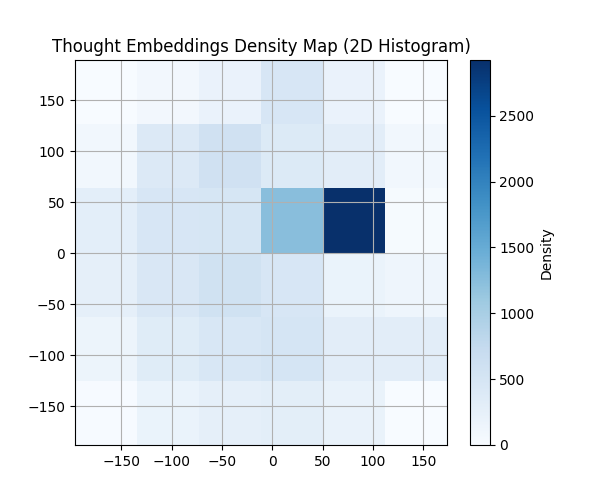}
        \label{fig:subheatmap2}
    \end{subfigure}
    \caption{The heatmap at the left represents the sampling before instruction evolution, while the one at the right represents the sampling after instruction evolution.}
    \label{fig:heatmap}
\end{figure}

\section{Experiment Setup}\label{appendix:setup}
\noindent \textbf{Models.} 
Unlike \textsc{AITZ}, we do not compare the CoaT result with the expected page and decide whether to roll back because most actions in real-device scenarios cannot be rolled back without cost. 
% Moreover, in the low-level evaluation of AndroidControl, we provide the low-level instruction to the agent as a CoaT Action. 
Previous work conducted continual pretraining on Qwen2-VL-7B using GUI domain data, resulting in a stronger base model. In our ablation study, we discuss the impact of continuous pretraining on IPL.
\citep{}

\noindent \textbf{Setup.} 
% \noindent \textbf{Hyperparameters.}
% All experiments were conducted on NVIDIA A100 GPUs. 
We conduct hyperparameter searches on AITZ to reproduce the baseline results and find that the optimal learning rate ranges from 3e-5 to 3e-6. 
Therefore, all baseline fine-tuning experiments adopt this setting.
Before IPL, during the instruction evolution stage, we apply LoRA fine-tuning with a LoRA rank of 128.
For IPL Stage 1, we use a learning rate between 1e-6 and 1e-7. In subsequent stages, we apply a constant learning rate of 1e-7. 
The batch size is consistently set to 128.
During fine-tuning (including baseline fine-tuning), we enable ViT training, whereas in the IPL phase, we experiment with freezing ViT.
For AITZ training, we followed the Falcons' approach, utilizing a maximum 1540×1540 resolution. For other experiments, we reduce the resolution to 1280×720 to optimize computational efficiency. The maximum context length is set to 32K for all experiments.
The fine-tuning experiments are conducted for 2 epochs, while IPL training is performed for 1 epoch. 
Since the large volume of Android control data, we sample 1/5 of the dataset for each IPL training iteration.

\noindent \textbf{CoaT Multi-turn Dialogue Prompts.}
\begin{enumerate}
    \item \textbf{Page Description.}
    \textit{Based on the mobile screenshot: Image URL, identify and describe the key elements visible on the screen, including any text, buttons, icons, input fields, or other interactive components.} 
    \item \textbf{CoaT Action Thought.} 
    \textit{Given the task: instruction, and considering the contextual details from the image alongside the full history of previous actions: action history, determine the most logical and effective next step. Focus on providing a clear, actionable, and goal-oriented response to advance the task.}
    \item \textbf{CoaT Action Description.}
    \textit{Task: Determine the Most Appropriate Next Step. Based on the previous analysis and the objective, determine the most appropriate next step to achieve the goal. Choose from the following options: - **click**: Select a button or specific UI element by specifying it clearly (e.g., `click xxx', where `xxx' is the button name or identifier).  
    - **scroll**: Perform a scrolling action if the required element is not visible, specifying the direction (e.g., `scroll up', `scroll down').  
    - **type**: Input specific text into a field or search bar, specifying the text clearly (e.g., type ``content'').  
    - **press**: Interact with device-level buttons such as Home, Back, or Enter, specifying the button (e.g., ``press Back'').  
    - **stop**: Conclude the task, indicating that the objective has been achieved. Provide the chosen action in the specified format and ensure it aligns with the analysis and the visible UI elements.}
    \item \textbf{Click Action Grounding.}
    \textit{As discussed earlier, your task now is to identify the precise screen region coordinates to tap for the action coat action. The coordinates must be integers and strictly within the range of 0 to 1000 for both axes. Please provide your response in the required format: <|box\_start|>(top\_x, top\_y),(bottom\_x, bottom\_y)<|box\_end|>. Ensure your output adheres to these constraints and remains concise.}
\end{enumerate}

\noindent \textbf{Instruction evolution Prompts.}
\begin{enumerate}
    \item \textbf{Page Description Annotation.}
    \textit{ I will provide you with a mobile page. Please describe the current page. Your description should include the content of the page and its general functionality. Please note that the descriptions you generate should be of moderate length. Your page description should match the actual image.} 
    \item \textbf{Action Thought Annotation.}
    \textit{**QUERY**: task,
    **ACTION HISTORY**: To proceed with the query, your past actions include: action history,
    **NEXT ACTION**: This is the next action you need to take: coat action,
    **TASK**: Given the screen and the above information, you have three tasks to do. First, you have to analyze what you have done. Second, you should analyze the screen for relevant details that might
    pertain to the given query. This includes checking for specific applications, icons, or buttons that are visible and any information or results that are currently
    displayed on the screen. 
    Tip: If the screen does not have the information you need, you can scroll left or scroll up to try to get the information.  Don't answer this logic question by saying that because the provided **NEXT ACTION** is..., therefore, the next action is... You need to think carefully on your own.
    You must answer the question with suitable lengths and the following format: 'Think: I have done..., Current screen is..., I need to... So the next action is ...' Your final action should be the same as the NEXT ACTION above.
    }
    \item \textbf{Q\&A Annotation.}
    \textit{Your goal is to draw inspiration from the given images and image description information to create multiple new questions and answers. This new creation is closely related to the given image and information, but the answers involved should be directly derived from the given information, because UI positions and UI text are one-to-one correspondence.
    Specifically, you should construct the following three types of questions and answers, a total of 15: 1. the function of some elements in the image. 2. Grounding questions and answers (the coordinates and approximate location of the target in the image). 3. Partial detailed information questions and answers (the structural relationship between multiple elements, type, style, etc.).
    Please try to keep your questions and answers diverse and informative, and ignore the message in the device status bar. 
    Here is the information related to the image: 
    UI positions: \{ui positions\},
    UI text: \{ui text\},
    coat screen desc: \{coat screen desc\},
    Please provide the following information in JSON format with the key questions and answers, and Don't add annotation parsing:}
\end{enumerate}
\begin{figure*}[ht]
  \centering
  \includegraphics[width=0.98\textwidth]{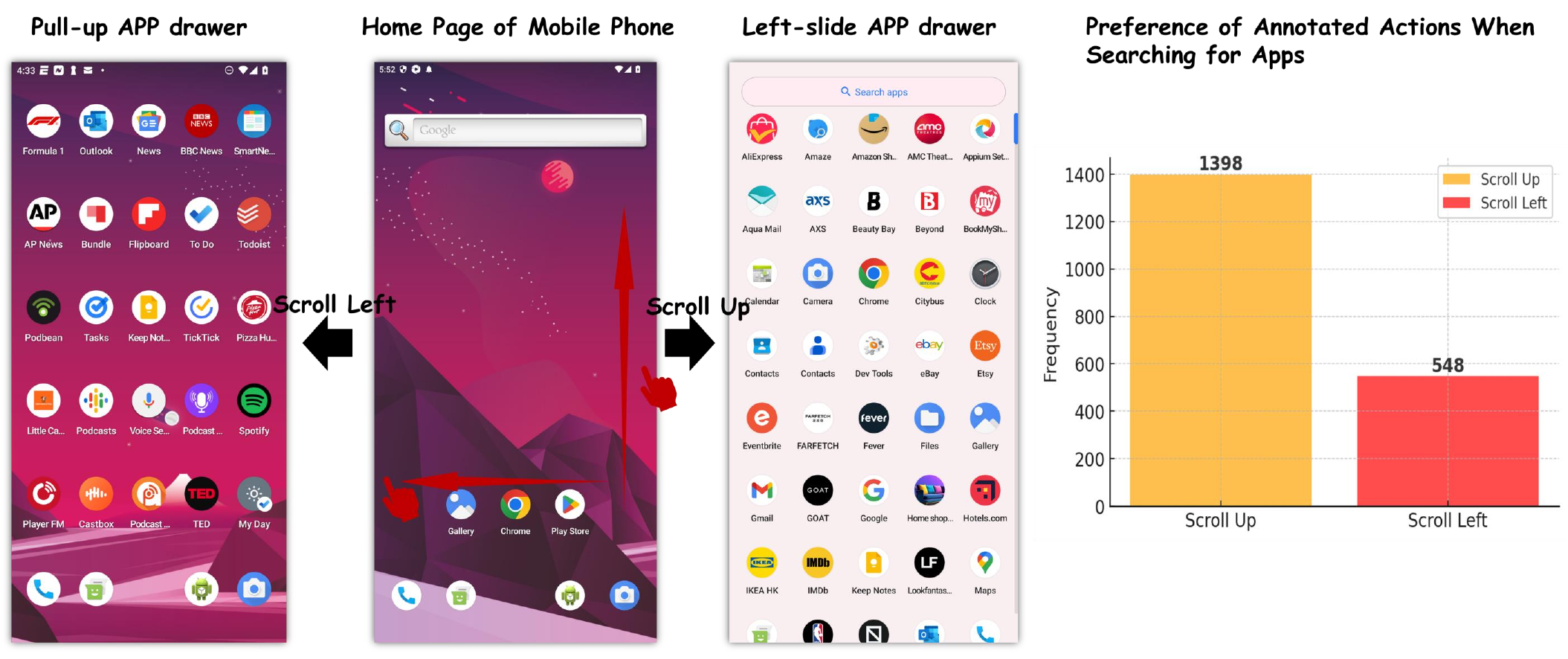}
  \caption{The left figure shows an example of unstable annotation preferences in \textbf{AMEX}, while the right figure presents the proportion of this type of annotation.}
  \label{fig:action_prefer}
\end{figure*}

\section{Human Annotation}
\noindent \textbf{SoTA Model Cost}: We use GPT-4o for annotation, which is priced at 4 USD per 1M input tokens and 16 USD per 1M output tokens. As shown in Appendix B (Instruction Evolution Prompts), each image has a resolution of 1080 × 2440. The input prompt is approximately 2K tokens, and the output is around 0.5K tokens, resulting in a per-page annotation cost of 0.008 USD (input) + 0.004 USD (output) = 0.012 USD. Given that the Android in the Zoo dataset [1] contains around 18,000 pages, the total annotation cost is approximately 200 USD.

\noindent \textbf{Human Verification}: Human verification is used to identify cases where the model produces incorrect thoughts due to being forced to align with the golden action. As shown in following table, some initial annotations contain template-like phrases (e.g., “Since”, “annotation”) and mention multiple possible actions. These mistakes often happen when the model tries to justify a given action even if it doesn’t match the actual screen.

The bolded text in following Table highlights common errors. To address this, we regenerated the prompts and applied automatic filtering based on specific tokens, which helps reduce the need for manual work.

\noindent \textbf{Before Human Verification:} "coat\_action\_think": "Reason for the Action:  - To find and compare prices, the user needs to locate similar products.  - Swiping up or scrolling down is a logical step to explore further information or discover additional options, as apps like Etsy often display "similar items" or "recommended products" below the primary product listing.  - This action aligns with typical app design patterns where related items or comparison options are placed further down the page. 4. Logical Derivation of Action: - Since the task specifies comparing prices of similar items, and the current screen does not explicitly list such items, the user must navigate downward to uncover more options.  - The instruction to "swipe up" (or scroll down) matches the action type ("action\_type": "scroll", "direction""down"), indicating an attempt to reveal additional content or sections. 5. \textbf{Since the next action swiping down matches the annotation, the next action is swiping down}.

\noindent \textbf{After Human Verification:}
Think: I have previously navigated "Back," which indicates that I may have returned to the previous screen or listing, likely as part of exploring options related to the query for comparing prices of men's athletic shoes. Current screen is displaying detailed information about a product on the Etsy app. This includes item details such as "Handmade item," "Made to order," and shipping information. There are also options to "Follow shop," "Message seller," and "Add to cart," but it does not appear to directly show other comparable items or prices for similar men's athletic shoes.I need to scroll further down on this screen to potentially uncover additional details, such as related products, a comparison section, or other useful information for the query. So the Next action is Scroll down."

% \section{IPL Scaling Analysis}\label{dpo scaling} 
% In the first four rounds of DPO pairs, the number of CLICK positive samples is 1731, 1496, 1485, 1475, while the number of PRESS positive samples is 580, 749, 579, 315.
% During the first round of IPL, when PRESS accuracy dropped to 10.18\%, CLICK has the highest count at 1731. In the following round, CLICK decreased by 235 (1731 → 1496), while PRESS increased by 169 (580 → 749). 
% After this round of training, PRESS Acc. recovers to 31.07\%, leading to a gradual decline in the proportion of this action type in the following round of sampling.
% In subsequent rounds, their ratio gradually stabilizes. For detailed sampling data changes, please refer to Figure \ref{fig:porpotion_click}.
\section{Iterative Sampling in the CoaT-tree}\label{appendix:sampling}
As shown in Figure \ref{fig:heatmap}, before instruction evolution, the distribution is highly concentrated, with only 8 points exceeding 1000 (including 3 points above 1200). After instruction evolution, the distribution becomes more balanced, with 20 points exceeding 1000 (including 2 points above 1500). 
\begin{figure*}[ht]
  \centering
  \includegraphics[width=0.98\textwidth]{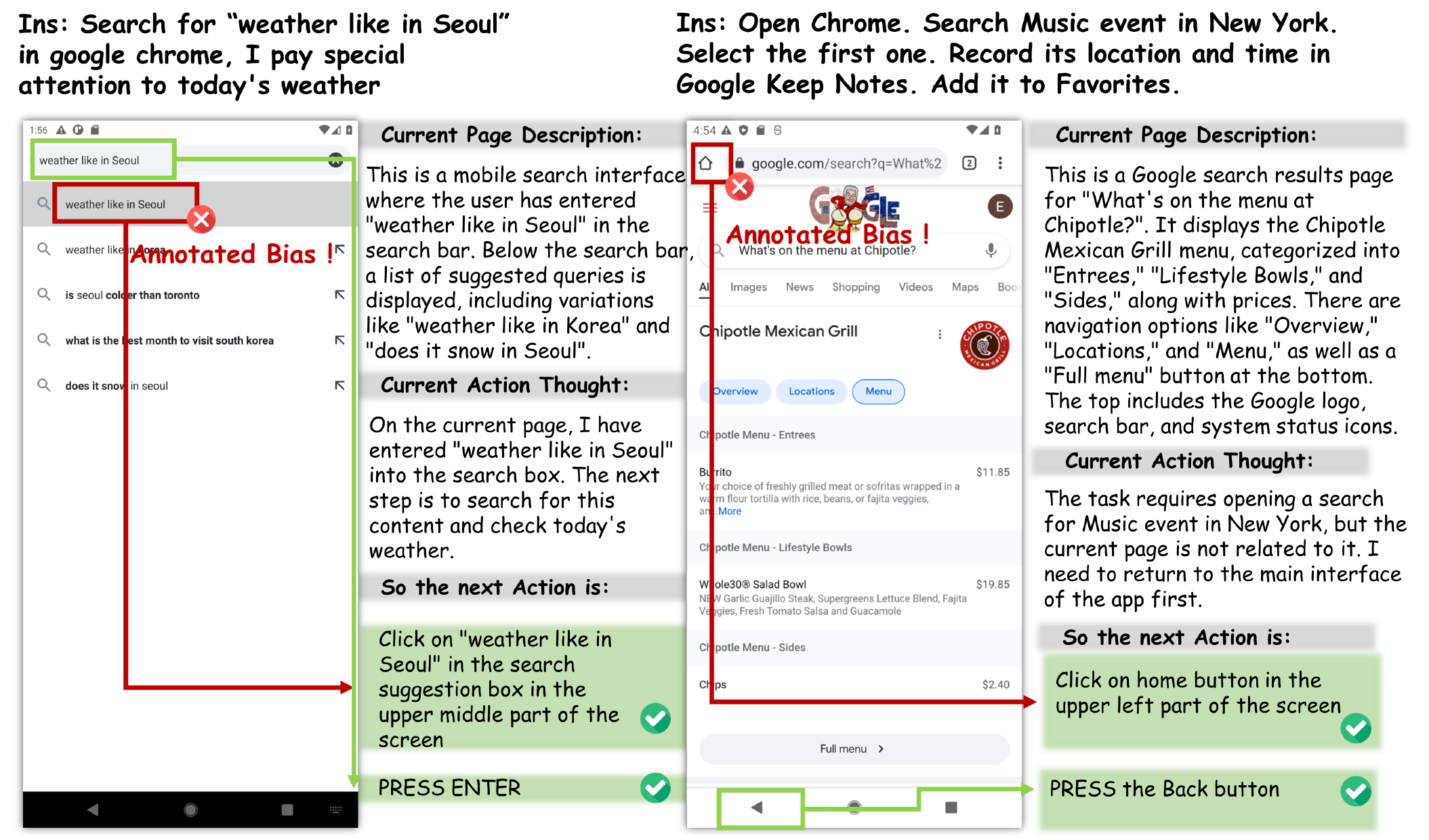}
  \caption{An example from \textbf{\textsc{AITZ}} demonstrates that when the task and image are the same, multiple actions may navigate to the same page.}
  \label{fig:annotation_bias}
\end{figure*}

\begin{figure*}[ht]
  \centering
  \includegraphics[width=0.98\textwidth]{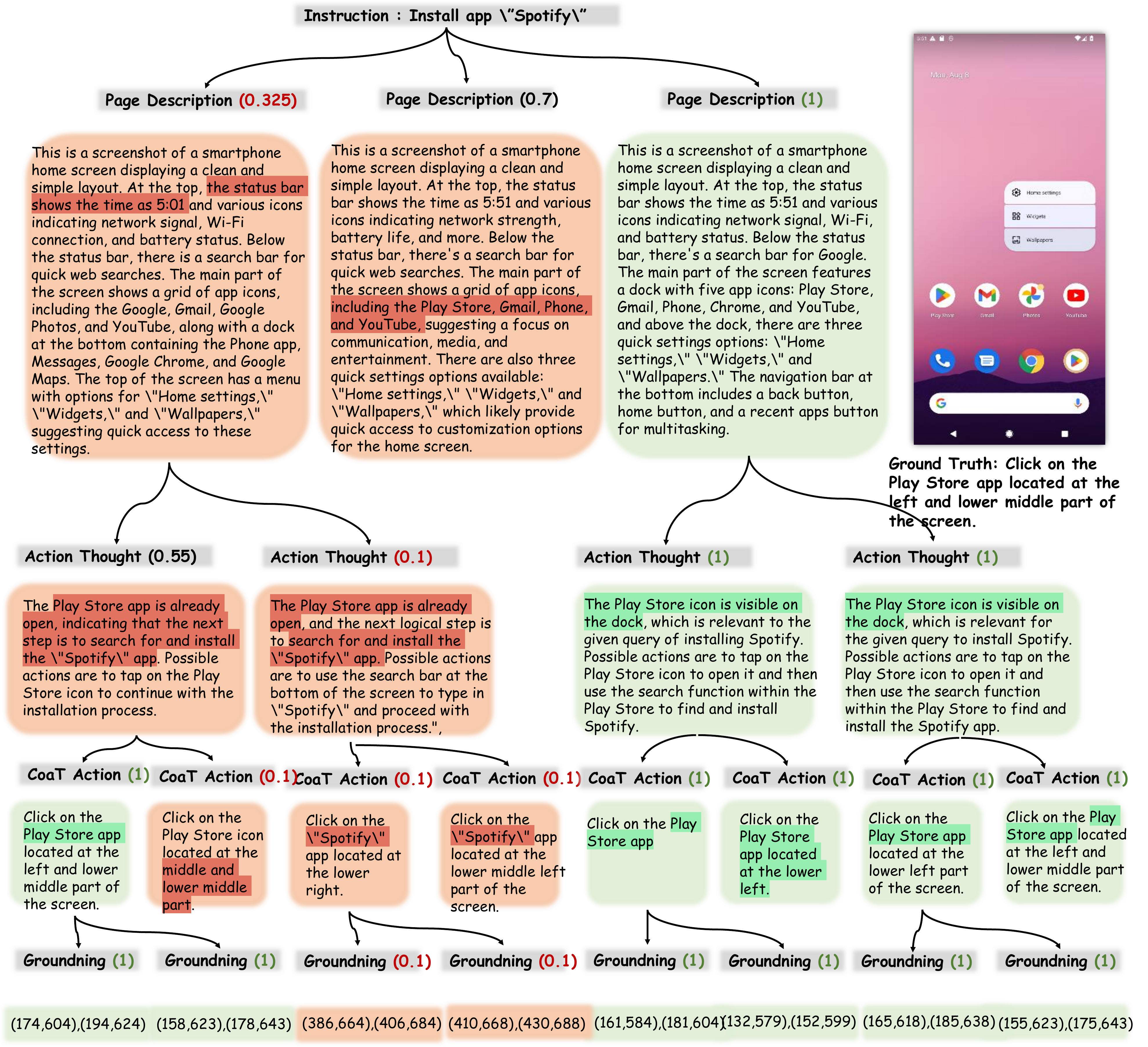}
  \caption{A sampling tree from \textbf{\textsc{AITZ}} demonstrates how the value is calculated.}
  \label{fig:sampling-case}
\end{figure*} 

\noindent \textbf{Potential correct space ratio.}
The proportion of $|\alpha| + |\beta|$ represents the potential correct space on the training data, and the change of this metric can clearly express the agent's ability to repair and reason out the correct process based on the correct answer. 

\section{Computational Cost of IPL Across Iterations}\label{compute_use}

\begin{table}[t]
\centering
\caption{Data size and training time across iterative rounds.}
\begin{tabular}{ccccccc}
\toprule
 & Standard SFT & Round 1 & Round 2 & Round 3 & Round 4 & Round 5 \\
\midrule
Data Samples & 63{,}158 & 156{,}418 & 34{,}010 & 28{,}491 & 17{,}027 & 17{,}780 \\
Time (hours)   & 2.8      & 7.9       & 2.0      & 1.7      & 1.0      & 1.0 \\
\bottomrule
\end{tabular}
\label{tab:iter-stats}
\end{table}

As shown in Table \ref{tab:iter-stats}, while the first IPL iteration incurs higher cost due to the larger volume of preference data, subsequent iterations are significantly lighter. The training time per round decreases rapidly, as the model generates fewer low-quality samples and requires fewer updates. In fact, by the third iteration, the training time becomes comparable to or even lower than the initial supervised fine-tuning (SFT) stage.

Therefore, the overall compute overhead of IPL remains moderate and manageable, especially considering its performance gains. Compared to SFT, IPL introduces only a modest increase in compute, but brings substantial improvements in reasoning and generalization.

\section{Case Study}\label{casestudy}
\noindent \textbf{Unstable annotation preferences.}\label{annotation preferences}
As shown in Figure \ref{fig:action_prefer}, the left section illustrates two different annotation preferences when searching for an app from the Home Page: \textbf{SCROLL UP} and \textbf{SCROLL LEFT}, leading to different destination pages.
The right part shows the overall preference distribution when annotators need to find an app. In rare cases, the annotation involves clicking on Google Play Store to perform a search. 
This phenomenon is quite common because, fundamentally, the task completion paths for a UI Agent are diverse. This is also the key difference between online evaluation and offline data evaluation.
From this, we observe that RL training on data with unstable preferences performs worse than SFT (e.g., \textsc{AITZ} SCROLL). 
This is because the DPO pair training method inherently attempts to correct errors in sampled preferences. 
As a result, the agent oscillates between two decisions when encountering the same GUI and instruction, failing to achieve consistent alignment.

\noindent \textbf{Action Equivalence.} 
Unlike Unstable Annotation Preferences, where different actions lead to different but equivalent pages, the issue here arises from annotators' random labeling habits in the training data, preventing the model from learning a consistent preference. 
Action Equivalence refers to the phenomenon where multiple actions on the same page can lead to the target page. 
However, since only one action is annotated as correct, other valid actions are mistakenly treated as incorrect.
As shown in Figure \ref{fig:annotation_bias}, after entering a search query, clicking on a suggested item in the recommendation bar, and pressing the Enter key on the keyboard produce the same effect. Similarly, when navigating back, clicking the on-screen back button and pressing the hardware back button yield the same outcome. 

\noindent \textbf{Thinking-Level Sampling.} 
As shown in Figure \ref{fig:sampling-case}, unlike mathematical reasoning, the CoaT process may not exhibit clear logical or computational errors. 
For a given action, a sampling CoaT data may produce hallucinations (Page Description) due to insufficient detail in the page description or fabricated elements; generate repetitive thoughts (Action Thought) due to neglecting action history; describe the wrong relative position of the correct element (CoaT Action); or misgrounding an element (Grounding), which is then classified as a negative sample. 
At the same time, outputs with more detailed and accurate descriptions, diversified thoughts, and different ways of describing the same widget are classified as positive samples.
Negative examples may be disadvantageous compared to positive examples, for example, because the description of the page is not detailed enough or the positioning of the elements is not accurate enough. 
At the same time, the wrong process may also give the correct result, but this is a very rare case. 
In this example, negative samples are generated due to the following three reasons: (1) \textbf{Rough page description: } The page contains eight app icons, but the agent's description includes only four apps: Play Store, Gmail, Phone, and YouTube; (2) \textbf{Hallucinated Thought:} The agent is unclear about its current page location. In reality, it is on the Home page, but it mistakenly believes it is in the Play Store (e.g., ''The Play Store app is already open"). (3) \textbf{Fabricated Position and Elements:} The agent generates the action "Click on the 'Spotify' app", even though there is no Spotify icon on the current page. This hallucination may stem from the instruction. 
Additionally, the Play Store icon should be located at the lower left part of the screen, but the agent incorrectly describes it as being in the middle and lower middle part.

\section{Usage of LLM statement } 
This paper utilized an LLM to improve the clarity and fluency of the text.

\end{document}